%% file: main.tex
\documentclass[10pt,journal,compsoc]{IEEEtran}
\input{preamble}
\begin{document}

\title{Contrastive-Augmented Flow Matching for\\
 Style-Content Disentanglement}

\author{Yusong Li*, ~\IEEEmembership{}
        Pingchuan Ma*,~\IEEEmembership{}
        Ming Gui,~\IEEEmembership{}
       Vincent Tao Hu,~\IEEEmembership{}
        Björn Ommer~\IEEEmembership{} %

\IEEEcompsocitemizethanks{\IEEEcompsocthanksitem * denotes equal contribution.

\IEEEcompsocthanksitem All authors are with the Computer Vision \& Learning Group at the University of Munich during this work.
E-mail: firstname.surname@lmu.de (all authors except Yusong Li); Yusong Li: liyusong1998@gmail.com.}
}

\maketitle

\input{Sec/abstract}

\input{Sec/Intro}

\input{Sec/Related}
\input{Sec/Method}
\input{Sec/Exp_new}

\input{Sec/Conclusion}

{\small
\bibliographystyle{IEEEtran}
\bibliography{main}
}

\clearpage
\input{Sec/Supp}

\end{document}

%% file: preamble.tex
\usepackage{amsmath,amsfonts}
\usepackage{algorithmic}
\usepackage{algorithm}
\usepackage{array}
\usepackage{textcomp}
\usepackage{stfloats}
\usepackage{url}
\usepackage{graphicx}
\usepackage{cite}
\usepackage{xcolor}
\usepackage{amssymb}
\usepackage{booktabs}
\usepackage{float}
\usepackage[accsupp]{axessibility}
\usepackage{multirow}
\usepackage{cleveref}
\usepackage{caption}
\usepackage[utf8]{inputenc}
\usepackage[T1]{fontenc}
\usepackage{microtype}
\usepackage{colortbl}

\definecolor{c1}{HTML}{765D97} 
\definecolor{c2}{HTML}{7BA67D}
\definecolor{c3}{HTML}{fc6160}
\definecolor{myblue}{HTML}{E6F3FC} 
\definecolor{mygray}{HTML}{DBE2E9} 
\definecolor{mygreen}{HTML}{006400}

\usepackage{xspace}

\makeatletter
\DeclareRobustCommand\onedot{\futurelet\@let@token\@onedot}
\def\@onedot{\ifx\@let@token.\else.\null\fi\xspace}

\def\eg{\emph{e.g}\onedot} 
\def\ie{\emph{i.e}\onedot}

\makeatother

%% file: Sec/abstract.tex
\begin{abstract}

Learning representations that separate content and style is crucial for controllable generation and compositional generalization. However, diffusion and flow-based models trained primarily with generative objectives often produce entangled or misaligned factors.
To address this gap, we introduce \textbf{C}ontrastive \textbf{A}ugmen\textbf{t}ed \textbf{F}low \textbf{M}atching (\emph{CAtFM}), a framework that integrates contrastive regularization into an invertible flow matching formulation to promote structured content–style representations.
Rather than constraining intermediate latents or velocity fields, we apply contrastive supervision to predicted endpoints during training, enforcing semantic consistency across transported distributions while allowing disentanglement to emerge implicitly, without assuming strictly pure or fully factorized content and style representations.
Our main experiments operate in CLIP embedding space, with additional validation using frozen DINO and ALIGN encoders. Across synthetic data, in-domain styles, and real-world benchmarks (ImageNet, WikiArt, DomainNet, and DTD), \emph{CAtFM} improves content and style retrieval, enhances embedding cluster separation, and achieves stronger open-set robustness compared to generative and discriminative baselines.
Overall, \emph{CAtFM} provides a simple way to couple discriminative constraints with deterministic transport, improving disentanglement and robustness under distribution shift. Code available at: \url{https://github.com/CompVis/SCFlow}
\end{abstract}


%% file: Sec/Intro.tex
\section{Introduction}
\label{sec:Intro}

Disentanglement is a central problem in representation learning and generative modeling, aiming to decompose underlying factors of variation into semantically meaningful variables~\cite{wang2024disentangledrepresentationlearning,Qi_2024,wang2021selfsupervisedlearningdisentangledgroup,frenkel2024implicit,shah2024ziplora,somepalli2024CSD}. In vision, this often corresponds to separating content and style into distinct latent or conditional representations~\cite{lachapelle2023synergies}, improving explainability, controllability, and generalization. From this perspective, disentanglement can be viewed as approximately inverting a structured image generator to recover the latent ``slots'' that generated an image, rather than merely predicting task-specific labels. A key desideratum is therefore \emph{compositional generalization}: the ability to handle out-of-domain scenes containing novel combinations of familiar factors (e.g., unseen content–style pairings) after observing each factor only in limited contexts.

\begin{figure}[!t]
\centering
\includegraphics[width=\columnwidth]{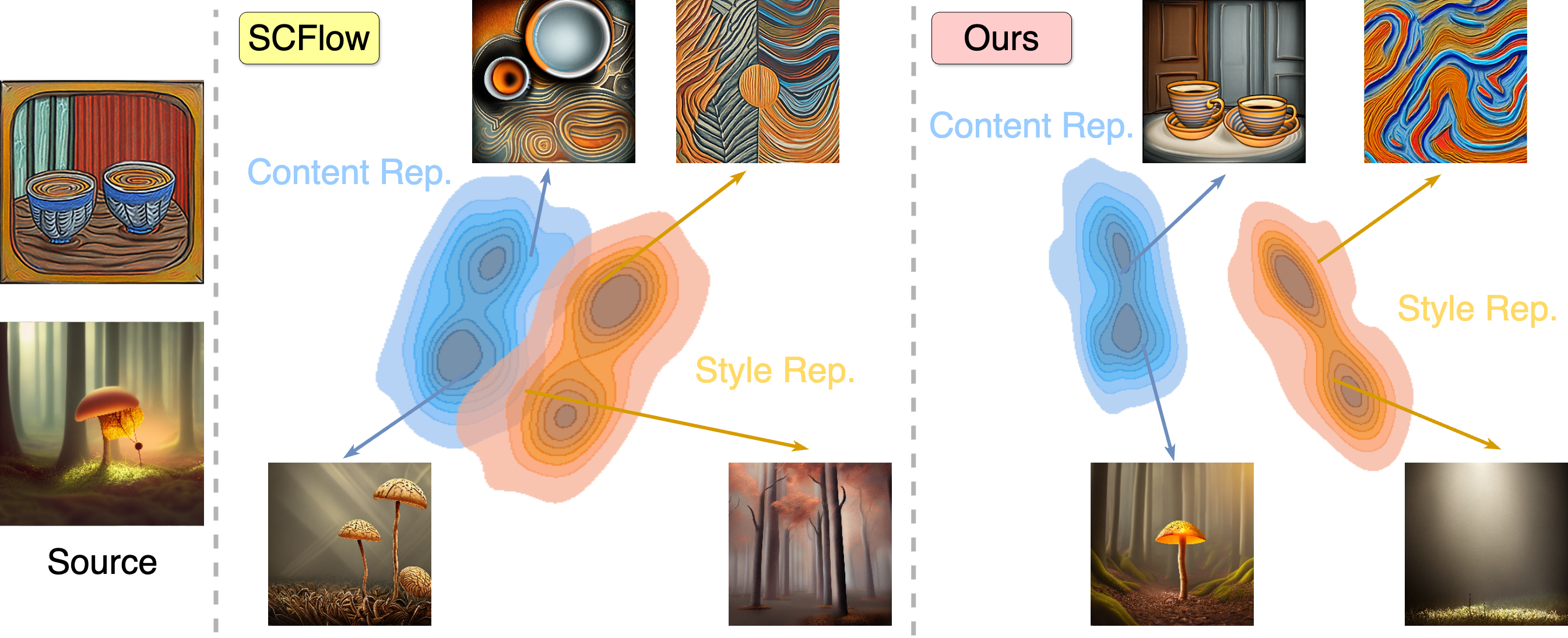}
\caption{Without explicit discriminative constraints, reconstruction-based generative training (\eg, SCFlow) fails to prevent content–style entanglement, particularly on unseen samples. (Zoom in for details.)}
\label{fig-leakage}
\end{figure}

Discriminative approaches address disentanglement by optimizing feature separability and decision boundaries~\cite{mo2023RepDisentangWithContraLearning,kahana2022contrastive,matthes2023towards}. Contrastive learning encourages representations in which samples sharing the same factor are close while others are pushed apart, often incorporating domain-wise constraints to enforce invariance to irrelevant variations~\cite{kahana2022contrastive}. Such methods can yield robust, task-relevant features and strong generalization when properly regularized~\cite{raina2003classification,xue2022investigating}.
However, they model only the conditional distribution $P(y \mid x)$ rather than the joint distribution $P(x,y)$. Consequently, they often struggle with compositional generalization to novel factor combinations~\cite{matthes2023towards,montero2021role} and cannot generate samples consistent with the underlying data structure. This limitation becomes more pronounced when $y$ itself decomposes into interacting factors (e.g., content and style) that jointly determine $x$.

On the generative side, many approaches enable controllable image synthesis by conditioning diffusion or flow-based models on textual or visual references for style transfer, editing, or domain adaptation~\cite{rout2024semantic,brack2024ledits++,li2024styletokenizer,li2023blipdiffusion,Qi_2024,xing2024csgo,frenkel2024implicit,gandikota2025sliderspace}. While these models produce high-quality images under rich conditioning, they typically lack explicitly disentangled latent variables for content and style, limiting direct factor-level analysis and manipulation. SCFlow~\cite{ma2025scflow} addresses this by learning to extract and recombine content and style without requiring ground-truth pure factors, reflecting realistic scenarios where fully factorized supervision is unavailable~\cite{ma2025scflow,zhang2025style}. This yields more interpretable embeddings for downstream tasks. However, because SCFlow relies solely on a generative objective, it exhibits \emph{unconstrained disentanglement}: content and style representations can leak into each other, resulting in semantically misaligned or incomplete factors (see~\Cref{fig-leakage}). More broadly, generative models trained without explicit discriminative constraints often struggle with combinatorial generalization~\cite{montero2021role}, \ie, are unable to handle novel combinations of familiar factors. These limitations motivate integrating contrastive objectives into flow-based transport, combining generative modeling with explicit semantic alignment.

\textit{Taken together, neither paradigm alone satisfies our objective.} Generative flow models such as SCFlow provide an invertible mechanism for blending and separation, yet reconstruction-only training permits content–style leakage. Discriminative methods enforce sharp factor boundaries but do not model the joint data manifold and therefore cannot support generation. SCFlow avoids explicit disentanglement supervision by learning an invertible merging process in latent space, where blending is defined through synthetic style–content pairs and separation arises from flow invertibility~\cite{ma2025scflow}. This motivates a simple remedy: retain the invertible flow while incorporating contrastive objectives that promote intra-factor compactness and inter-factor separation, yielding compositionally stable embeddings without sacrificing the invertibility.

Concretely, we propose \emph{CAtFM}, an extension of SCFlow that incorporates contrastive supervision that reconstructs source and target endpoints from the learned velocity field and applies contrastive objectives to representations derived from the transport process. This unifies generative flow transport with discriminative regularization, mitigating factor leakage observed under generative training while preserving SCFlow’s implicit, supervision-light formulation. Conceptually, the contrastive term acts as an inductive bias on the latent generator, encouraging more semantically consistent and less interfering content–style slots even under unseen compositions, analogous to structural constraints on decoder classes for compositional perception~\cite{brady2025generation}.
Empirically, across SCFlow’s synthetic dataset, in-domain unseen styles, and real-world benchmarks (e.g., ImageNet~\cite{deng2009imagenet} and WikiArt~\cite{saleh2015large}), our method improves style purity, content fidelity, clustering structure, and retrieval performance, demonstrating that invertible latent generators provide a strong scaffold for disentanglement but benefit from explicit contrastive constraints to avoid factor leakage.

%% file: Sec/Related.tex
\section{Related Work}
\label{sec:related_work}

\subsection{Representation Disentanglement}
\label{disentanglement}
Disentanglement has long been a central objective in representation learning. 
Traditional approaches such as $\beta$-VAE~\cite{higgins2017betavae} and FactorVAE~\cite{kim2019disentanglingfactorising} pursue \textbf{axis-aligned disentanglement}, where each latent dimension corresponds to a single generative factor. Metrics including Mutual Information Gap (MIG)~\cite{chen2019isolatingsourcesdisentanglementvariational}, Separated Attribute Predictability (SAP)~\cite{kumar2018variationalinferencedisentangledlatent}, and Disentanglement, Completeness, Informativeness (DCI)~\cite{eastwood2018a} quantify alignment between latent variables and underlying factors. While effective on low-dimensional data with simple generative structures, these methods rely on factorized and axis-aligned representations. This breaks down for real-world artistic images, where style and content interact non-linearly and semantically~\cite{kotovenko2019iccv}. 

Moreover, unsupervised disentanglement is provably unidentifiable without strong inductive biases~\cite{locatello2019challengingcommonassumptionsunsupervised}, and axis-aligned metrics become unreliable in high-dimensional regimes. Moreover, ~\cite{montero2024lostlatentspacedisentangled} shows that even highly disentangled latent models fail to generalize to unseen combinations of factors, highlighting that pure factorization alone does not ensure semantic robustness. In contrast, we approach disentanglement from a generative distributional perspective, focusing on reconstruction and cross-distribution transfer rather than axis alignment. This aligns with recent work that reframes disentanglement as structured style–content modeling in complex visual domains.

\subsection{Style and Content Representation}
\label{style_content_representation}
Recent research on content and style representation has evolved along \textit{discriminative and generative approaches.} 
Discriminative methods learn structured embeddings through supervised or contrastive objectives. Early works classified visual style~\cite{karayev2013recognizing, saleh2015large}, while later approaches ~\cite{somepalli2024CSD, wang2023evaluating, li2024styletokenizer} employed contrastive learning on curated or synthetic data to build semantically consistent style descriptors. Large-scale contrastive models such as CLIP~\cite{radford2021learning} and DINO~\cite{dino,oquab2024dinov2learningrobustvisual} yield strong content representations but often preserve stylistic bias due to data and objective coupling.

In the generative direction, neural style transfer~\cite{gatys2016image} and its extensions~\cite{johnson2016perceptual, li2019learning, wang2023microast, wang2020glstylenet, kotovenko2021rethinking, zuo2022style, zhang2022domain} pioneered controllable blending of style and content through feature transformation. More recently, diffusion-based methods~\cite{li2023blipdiffusion, Qi_2024, xing2024csgo, frenkel2024implicit, gandikota2025sliderspace} enable reference-driven content or style injection, but typically do not enforce explicit semantic separation between these factors. SCFlow~\cite{ma2025scflow} formulates disentanglement via a purely generative objective with an asymmetric flow from source to target distributions, implicitly isolating content and style by progressively removing residual information. Although reverse flow inference improves representation quality, the learned embeddings remain unconstrained, allowing information leakage and semantic misalignment. Overall, existing generative and discriminative approaches lack mechanisms that explicitly align semantic structure while preserving distributional modeling. This limitation motivates our contrastive–flow formulation.

\subsection{Contrastive Learning}
\label{contrastive learning}
Contrastive learning originated with the contrastive loss~\cite{chopra2005learning} for face recognition and verification, where embeddings are trained to pull together samples of the same identity and push apart different ones~\cite{hadsell2006dimensionality,weinsberger2009lmnn}.
It later became a core paradigm for self-supervised learning, where positive and negative pairs are constructed from unlabeled data via data augmentations~\cite{chen2020simpleframeworkcontrastivelearning, infoNCE, he2020momentumcontrastunsupervisedvisual}.
In both supervised and self-supervised settings, the objective shapes the latent space through similarity-based compactness of positives and separation of negatives.

Existing methods can be broadly divided into two categories. Pair-based metric learning directly optimizes distances between embedding pairs to enforce class-level clustering~\cite{contrastiveloss,wu2018samplingmattersdeepembedding,wang2020multisimilaritylossgeneralpair}. Proxy-based metric learning introduces learnable representatives that approximate class structure, reducing computational cost while modeling global embedding geometry~\cite{movshovitzattias2017fussdistancemetriclearning, teh2020proxyncarevisitingrevitalizingproxy, kim2020proxyanchorlossdeep}. In proxy-based formulations, proxies are jointly optimized with network parameters and serve as anchors summarizing subsets of training data. Overall, the contrastive formulation aligns naturally with our objective of enforcing semantic consistency across representations transformed via flow matching.

\subsection{Flow-based models}
\label{diff and fm}

Diffusion models~\cite{sohl2015deep,ho2020denoising,song2020denoising_ddim, song2021scorebased_sde} formulate generation as learning to reverse a stochastic denoising process. A forward diffusion gradually perturbs data by adding Gaussian noise until it converges to an isotropic prior, while a reverse diffusion network is trained to denoise and recover samples from this prior. Inversion techniques such as DDIM inversion~\cite{song2020denoising_ddim, mokady2023null} and related SDE-based methods~\cite{meng2021sdedit, hertz2022prompt}, as well as Dual Diffusion Implicit Bridges (DDIB)~\cite{su2023dualdiffusionimplicitbridges} further enable bidirectional mapping between clean and noisy samples, allowing applications in image editing and conditional synthesis through controlled noise manipulation.

Flow Matching (FM)~\cite{lipman2023flowmatchinggenerativemodeling,rectifiedflow_iclr23,albergo2023stochastic,neklyudov2023action} generalizes diffusion by replacing stochastic noise injection with deterministic transport between source and target distributions, parameterized as an ordinary differential equation(ODE). Unlike diffusion, FM per mits flexible noise schedules and base distributions and does not require an isotropic Gaussian prior. This flexibility has enabled applications in low-to-high resolution translation~\cite{2023boosting}, image–depth translation~\cite{gui2024depthfm}, and text–image generation~\cite{liu2025flowing,he2025flowtok}.

Beyond pure generation, recent works explore discriminative FM variants to improve representation structure and semantic alignment. Some~\cite{stärk2024harmonicselfconditionedflowmatching,davis2024fisherflowmatchinggenerative} predict target samples directly rather than velocities, but still require velocity computation at inference, increasing overhead. $\Delta$-FM~\cite{stoica2025contrastiveflowmatching} and dispersive loss~\cite{wang2025diffusedisperseimagegeneration} introduce contrastive regularization to encourage class separation and coherent intermediate representations along the flow. These results suggest that discriminative constraints can significantly strengthen feature geometry and generalization in flow-based models.

In contrast, we propose a novel contrastive-augmented FM formulation that predicts both source and target endpoints from the learned velocity. Contrastive objectives are then applied to the predicted embeddings to explicitly enforce semantic alignment of content and style, while disentanglement itself remains implicitly learned through the generative flow dynamics rather than being directly supervised.

%% file: Sec/Method.tex
\section{Method}
\label{sec:method}

\begin{figure}
\centering
\includegraphics[width=1.0\columnwidth]{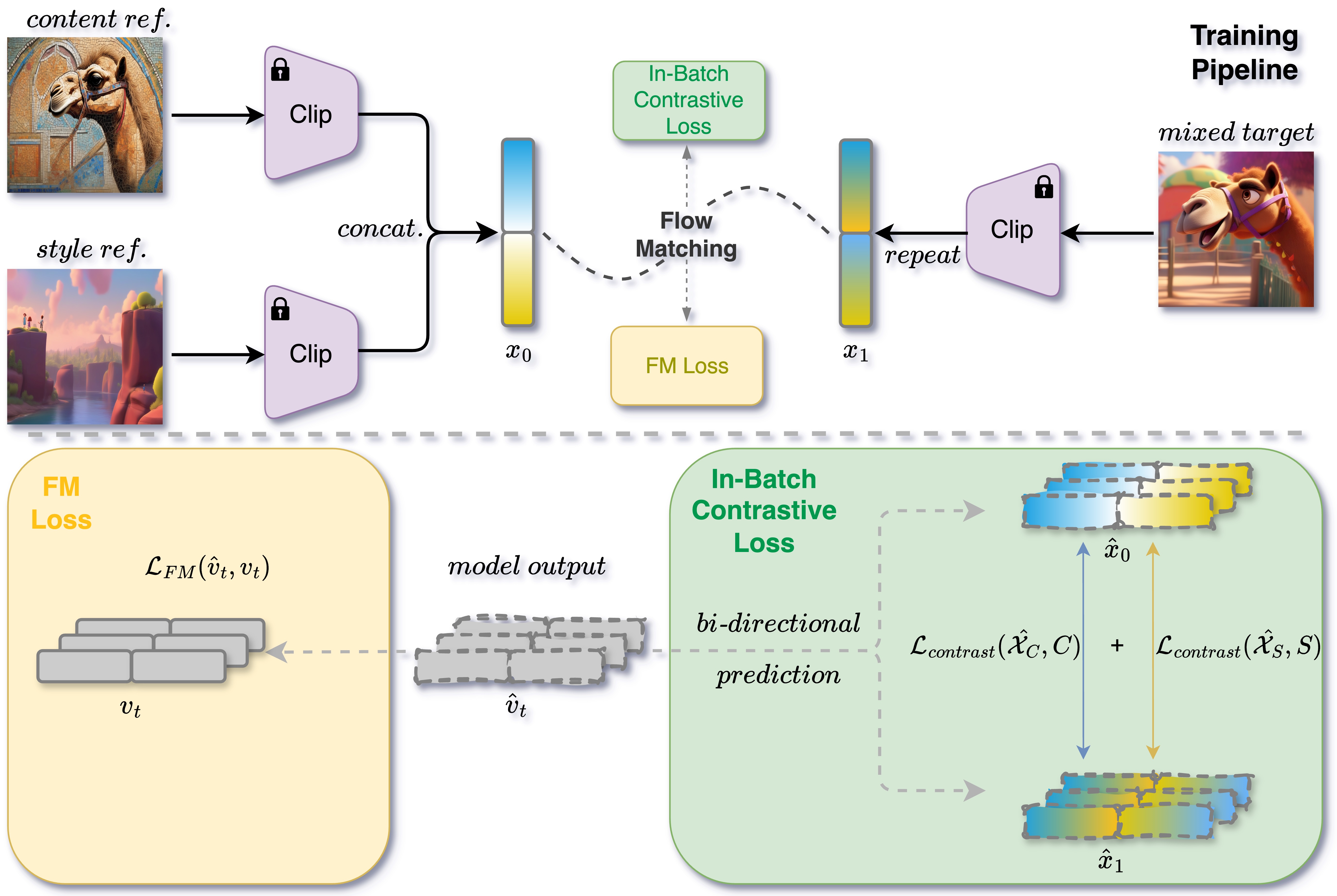}
\caption{Our Method.}
\label{fig:method}
\end{figure}

\emph{CAtFM} jointly models style–content blending and disentanglement as a deterministic transport problem under the Flow Matching (FM) framework~\cite{lipman2023flowmatchinggenerativemodeling}.
~\Cref{sec:method_preliminaries} revisits contrastive learning and the FM formulation;
~\Cref{sec:fm_endpoints} details the construction of source–target endpoint pairs from content–style triplets.
~\Cref{sec:method_main} presents the complete training procedure, combining bidirectional endpoint prediction with contrastive regularization.
The full algorithm is summarized in~\Cref{alg:alg1}. ~\Cref{fig:method} illustrates the overall training pipeline.

\subsection{Preliminaries}
\label{sec:method_preliminaries}
\subsubsection{Contrastive Learning}
\label{sec:method_pre_contrastive_learning}

Contrastive learning aims to learn an embedding function that maps input into a space where distances reflect semantic similarity. Given a batch of $B$ samples $\mathcal{X} = \{x_1, x_2, \ldots, x_B\}$ with associated labels or semantic contexts $\mathcal{Y}$, the objective encourages embeddings of semantically similar samples to cluster together while separating dissimilar ones. For each anchor $x_i$, positive samples $x_i^+$ share the same semantic context, whereas negatives $x_j$ correspond to different classes or contexts defined by $\mathcal{Y}$

A common formulation is the pair-based objective, which optimizes embeddings similarity using sampled positive and negative pairs within the batch. Among these, InfoNCE loss~\cite{infoNCE} is widely adopted and can be written as~\cite{wang2025diffusedisperseimagegeneration}:

\begin{equation}\label{eq:infoNCE}
\mathcal{L}_{\text{InfoNCE}} = D(x_i, x_i^+) + \log \sum_{j=1}^B -D(x_i, x_j),
\end{equation}
where $D(\cdot, \cdot)$ denotes a distance function, typically implemented as a scaled cosine similarity. 
More generally, a contrastive learning objective can be written as:
\begin{equation}\label{eq:contrastive}
\mathcal{L}_{\text{contrast}}(\mathcal{X}, \mathcal{Y}),
\end{equation}
where $\mathcal{X}$ denotes the input samples and $\mathcal{Y}$ their labels or semantic contexts that define positive and negative relationships. We use this abstract notation throughout to cover different contrastive losses with varying sampling and optimization schemes, unless specified otherwise.

\subsubsection{Flow Matching}
\label{sec:method_pre_FM}
Flow Matching (FM) forms the foundation of our method, enabling deterministic transport between the disentangled distribution $p_0(x)$ and merged distribution $p_1(x)$. We define the time-dependent interpolation process~\cite{albergo2023stochastic} for $t \in [0,1]$:
\begin{equation}\label{eq:fm-interpolation}
x_t = \alpha_t x_0 + \sigma_t x_1,
\end{equation}
where $x_0$ denotes the source representation (content and style references) and $x_1$ represents the merged target. The interpolation coefficients $\alpha_t$ and $\sigma_t$ control the transition between endpoints, typically using a linear schedule $(\alpha_t=1-t, \sigma_t=t)$ that satisfy the boundary conditions $\alpha_0=\sigma_1=1$ and $\alpha_1=\sigma_0=0$. As $t$ evolves from 0 to 1, the trajectory smoothly transports samples from $p_0(x)$ to $p_1(x)$.

The trajectory dynamics are governed by a velocity field $v(x,t)$ satisfying the ordinary differential equation (ODE)
\begin{equation}
\frac{dx}{dt} = v(x,t), \quad
v(x,t) = \mathbb{E}[\dot{x}_t | x_t = x],
\end{equation}
which induces the intermediate marginal distribution $p_t(x)$ over time~\cite{lipman2023flowmatchinggenerativemodeling,ma2024sit,song2021scorebased_sde}. A neural network $v_\theta(x_t,t)$ is trained to approximate this velocity field by minimizing the flow-matching loss:
\begin{equation}
\label{eq:fm-loss}
\mathcal{L}_{\text{FM}}(\theta)
= \mathbb{E}_{t,x_0,x_1} \big[|v_\theta(x_t, t) - \dot{\alpha}_t x_0 - \dot{\sigma}_t x_1 |^2 \big].
\end{equation}

During inference, the learned velocity field is integrated using an ODE solver:

\begin{equation}
\text{ODESolve}(x_t, v_{\theta})_{[0,1]} = x_0 + \int_0^1 v_\theta(x_t, t) dt,
\end{equation}
to obtain the merged result $x_1$ from input $x_0$, and conversely recovers $x_0$ by integrating the ODE backward in time.

\subsection{Constructing Flow Matching Endpoints}
\label{sec:fm_endpoints}
We seek to learn a bidirectional mapping between disentangled and merged distributions, enabling content–style blending in the forward direction and separation in the reverse direction via a single flow field. A naive straightforward formulation would define $x_0 = (c, s)$ sampled from a disentangled distribution $p_0(x)$ and its stylized counterpart $x_1 = c \oplus s$ sampled from a merged distribution $p_1(x)$. However, explicit supervision of pure content and style factors $c, s$ is generally unavailable, as these attributes are inherently entangled in real images. Instead, we construct the FM starting point $x_0$ as a pair of content and style reference samples, each potentially containing extraneous style or content information not present in the blended target. This forms a triplet structure:
\begin{equation}
    (I_{c_is_*}, I_{c_*s_j}, I_{c_is_j})
\end{equation}
Here, $I_{c_i s_*}$ denotes an image with fixed content $c_i$ and a randomly sampled style, $I_{c_* s_j}$ represents an image with fixed style $s_j$ and random content, and $I_{c_i s_j}$ is the merged target image combining both $c_i$ and $s_j$. The symbol ``$*$'' indicates a marginalized factor sampled independently of the specified variable. 
The model learns a bidirectional mapping between the two distributions, performing blending in the forward direction $(p_0 \rightarrow p_1)$ and separation in the reverse direction $(p_1 \rightarrow p_0)$. Disentanglement arises implicitly from the invertibility of the generative transport rather than from explicit factor supervision. Owing to the deterministic and invertible nature of FM (see \Cref{sec:method_pre_FM}), training in a single direction suffices to enable transitions between arbitrary endpoint distributions, without imposing the Gaussian prior constraint typical of diffusion models~\cite{ldm, fuest2024diffusion}.

We operate in the latent space of a pretrained feature extractor (\eg CLIP~\cite{radford2021learning}), which provides semantically rich and compact representations.
When applicable, these embeddings can be visualized or decoded through generative backends such as unCLIP~\cite{ramesh2022hierarchical}.
Let $E(\cdot)$ denote a pretrained image encoder, which maps an image $I$ to its latent embedding $z = E(I)$.
These three latents are used to form the source–target pair for FM: %
\begin{align}
\label{eq:FM-endpoints}
    x_0 = [z_{c_i,s_*}, z_{c_*,s_j}]\sim p_0(x),\\
    x_1 = [z_{c_i,s_j}, z_{c_i,s_j}]\sim p_1(x),
\end{align}
where $x_0$ denotes the concatenation of the content and style references, possibly containing additional irrelevant information, while $x_1$ corresponds to the merged target, repeated to match dimensionality.

\subsection{\emph{CAtFM}: Contrastive-Augmented Flow Matching}

\noindent\textbf{Contrastive Guidance in Flow Matching.} 
Existing works incorporate contrastive learning into diffusion or flow matching models by applying the objective either to the predicted velocity field~\cite{stoica2025contrastiveflowmatching} or to intermediate network activations~\cite{wang2025diffusedisperseimagegeneration}.
In $\delta$-FM~\cite{stoica2025contrastiveflowmatching}, the contrastive term operates in velocity space, separating flows conditioned on different labels while the standard FM loss fits the ground-truth transport direction.. Similarly, Diffuse and Disperse~\cite{wang2025diffusedisperseimagegeneration} applies an InfoNCE-style objective to hidden features, acting as a global representation regularizer that shapes feature geometry without targeting a specific factorization such as content and style. These approaches implicitly assume that velocity fields or internal activations possess stable, class-consistent semantics, \eg, flows between a shared Gaussian prior and class-specific distributions, or features that encode a single dominant semantic label.

However, this assumption does not hold in our setting. The flow endpoints (\eg, CLIP embeddings) are not independent class distributions, but coupled content–style pairs whose semantics depend jointly on both inputs. As a result, the same style label combined with different content induces distinct velocity trajectories. Likewise, intermediate activations along the flow encode entangled content and style signals with input-dependent roles, making it unclear which layer or subspace can be consistently interpreted as ``content'' or ``style'' across samples. These observations motivate a principled alternative: rather than contrasting ambiguous internal quantities, we use the predicted velocity $v_{\theta}$ to reconstruct both FM endpoints: the source $\hat{x}_0$ and target $\hat{x}_1$, applying contrastive objectives directly to these reconstructed representations. 

Although this usually requires expensive iterative ODE integration during training, the compact latent space and linear schedule enable efficient one-step sampling:
\begin{align}
\label{eq:FM-bidirection-predict}
    \hat{x}_0 &= x_t - t\cdot v_\theta(x_t,t),\\
    \hat{x}_1 &= x_t + (1-t)\cdot v_\theta(x_t,t),
\end{align}
where the target velocity $v(x,t) = x_1 - x_0$ under the linear schedule $(\alpha_t = 1-t, \sigma_t = t)$~\cite{lipman2024flowmatchingguidecode}.
This design enables meaningful content- and style-level consistency, aligning discriminative and generative objectives while avoiding the instability that arises when contrastive supervision is applied to intermediate flow states or hidden features. 

Each prediction is formed by concatenating either the content and style reference (inputs) or two copies of the target embedding as defined in~\cref{eq:FM-endpoints}. Since the first half of the concatenated representation corresponds to content and the second to style, we apply in-batch contrastive losses~(\Cref{eq:contrastive}) independently to each half, using the respective content ($y_c$) and style ($y_s$) labels: 
\begin{align}
\label{eq:l_cont}
    \mathcal{L}_{cont
} = \mathcal{L}_{\text{contrast}}(\hat{\mathcal{X}_C}, \mathcal{C}) +  \mathcal{L}_{\text{contrast}}(\hat{\mathcal{X}_S}, \mathcal{S}) ,
\end{align}
where $\hat{\mathcal{X}}$ denotes the concatenation of $\hat{\mathcal{X}_0}$ and $\hat{\mathcal{X}_1}$ along batch dimension, $\mathcal{C}$ and $\mathcal{S}$ representing the corresponding content and style labels. Let $h$ denotes the length of a single vector $z$. The effective batch size for each contrastive objective is thus increased from $B$ to $2B$, enlarging the pool of in-batch negatives and strengthening discriminative learning~\cite{yeh2022decoupledcontrastivelearning,chen2020simpleframeworkcontrastivelearning,chen2022largerbatchsizeincontrastivelearning}. The choice of contrastive loss remains flexible, allowing different formulations to be adapted to specific scenarios. The overall training objective is defined as:
\begin{equation}
    \mathcal{L}_{\text{total}} = \mathcal{L}_{FM} + \lambda \cdot \mathcal{L}_{cont},
\end{equation}
where $\lambda$ denotes the weight assigned to the contrastive objectives. The full training procedure is summarized in~\Cref{alg:alg1}, and~\Cref{fig:method} illustrates the overall pipeline.

\label{sec:method_main}
\begin{algorithm}
\caption{ Training Algorithm.}
\label{alg:alg1}
\begin{algorithmic}[1]

\STATE {\textbf{Input}: 
Triplets $(z_{c_i s_*}, z_{c_* s_j}, z_{c_i s_j})$, content labels $\mathcal{C}$, style labels $\mathcal{S}$
}
\vspace{2pt}
\STATE {\textbf{Hyperparameter}: batch size $B$, contrastive loss weight $\lambda\in [0,1)$ }
\STATE {\textbf{for} each training step \textbf{do}:}
\vspace{2pt}
\STATE \hspace{0.2cm} {\textit{// construct flow-matching endpoints}}
\vspace{2pt}
\STATE \hspace{0.5cm} {$x_0 \gets \text{concat}(\,z_{c_i s_*},\, z_{c_* s_j}\,)$}
\vspace{2pt}
\STATE \hspace{0.5cm} {$x_1 \gets \text{concat}(\,z_{c_i s_j},\, z_{c_i s_j}\,)$}
\vspace{2pt}
\STATE \hspace{0.5cm} {$t \sim \text{Uniform}(0,1) $ }
\vspace{2pt}
\STATE \hspace{0.5cm} {$v_t \gets x_1 - x_0 $ }
\vspace{2pt}
\STATE \hspace{0.5cm} {$x_t \gets (1-t)x_0 + tx_1 $ }
\vspace{2pt}
\STATE \hspace{0.5cm} {$ \hat{v_t} \gets v_\theta(x_t, t) $}
\vspace{2pt}
\STATE \hspace{0.2cm} {\textit{// FM loss}}
\vspace{2pt}
\STATE \hspace{0.5cm} {$ \mathcal{L}_{\text{FM}} \gets \frac{1}{B}\sum_{n=1}^B ||\hat{v_t}^{(n)}-v_t^{(n)}||_2^2$}
\vspace{2pt}
\STATE \hspace{0.2cm} {\textit{// bidirectional prediction}}
\vspace{2pt}
\STATE \hspace{0.5cm} {$ \hat{x}_0 \gets x_t - t \hat{v_t}$}
\vspace{2pt}
\STATE \hspace{0.5cm} {$ \hat{x}_1 \gets x_t + (1-t) \hat{v_t}$}
\vspace{2pt}
\STATE \hspace{0.2cm} {\textit{// general in-batch contrastive loss follow~\Cref{eq:l_cont}}}
\vspace{2pt}
\STATE \hspace{0.5cm} {$\hat{\mathcal{X}} \gets \text{stack}(\hat{\mathcal{X}_0}, \hat{\mathcal{X}_1})$ }
\STATE \hspace{0.2cm} {\textit{// define content and style splits}}
\STATE \hspace{0.5cm} $\hat{\mathcal{X}_C} \gets \hat{\mathcal{X}}[:, :h]$  %
\STATE \hspace{0.5cm} $\hat{\mathcal{X}_S} \gets \hat{\mathcal{X}}[:, h+1:]$ %
\vspace{2pt}
\STATE \hspace{0.5cm} {$\mathcal{L}_{\text{cont}
}\gets \mathcal{L}_{\text{contrast}}(\hat{\mathcal{X}}_{C}, \mathcal{C})+ \mathcal{L}_{\text{contrast}}(\hat{\mathcal{X}}_{S},\mathcal{S})$}
\STATE \hspace{0.2cm} {\textit{// total loss}}
\vspace{2pt}
\STATE \hspace{0.5cm} {$\mathcal{L}_{\text{total}} \gets \mathcal{L}_{FM} + \lambda \cdot \mathcal{L}_{\text{cont}}$}
\vspace{2pt}
\STATE \hspace{0.5cm} {\textit{Update $\theta$ w.r.t. $\mathcal{L}_{\text{total}}$}}

\end{algorithmic}
\end{algorithm}

\noindent\textbf{Sampling.} For inference, we integrate the learned velocity field using an ODE solver to perform both forward merging and reverse separation, following SCFlow.
The forward merge is defined as:
\begin{equation}
\label{eq:forward}
z_{c_i, s_j} = \texttt{mean}\big(\text{ODESolve}([z_{c_i, s_*}, z_{c_*, s_j}], t_0=0, t_1=1)\big),
\end{equation}
where the $\texttt{mean}$ operator aggregates the two halves of the concatenated latent to produce a single embedding corresponding to the stylized output.
The reverse separation, given a single input embedding, is defined as
\begin{equation}
\label{eq:reverse}
[z_{c_i,\bar{s}}, z_{\bar{c}, s_j}] = \text{ODESolve}([z_{c_i, s_j},z_{c_i, s_j}], t_0=1,t_1=0).
\end{equation}
Here, $\bar{s}$ and $\bar{c}$ denote the average style and content components conditioned on $c_i$ and $s_j$, respectively, over the dataset. See the Supplementary Material for complete inference pipeline.

%% file: Sec/Exp_new.tex
\section{Experiments}
\label{sec:experiments}

We present both quantitative and qualitative evaluations of our approach. We compare its performance against contrastive baselines and state-of-the-art methods, and further analyze its generalization and robustness across different datasets and feature spaces. 

\subsection{Experimental Setup and Evaluation}
\label{sec:exp_setup_eval}

\subsubsection{Training and Implementation}
\label{sec:exp_training}
We follow the same architecture and training protocol across all methods to ensure a fair comparison.
All trained models use the 12-layer 1D Transformer backbone of SCFlow~\cite{ma2025scflow} and are trained for 120k steps with a learning rate of $1\mathrm{e}{-5}$.
We use batch size $B{=}384$ and set $\lambda{=}0.5$ as the weight of the in-batch contrastive loss. Default hyperparameters (\eg, temperatures and margins) are retained from the original works.
We find that using distinct objectives for content (InfoNCE~\cite{infoNCE}) and style (Multi-Similarity loss~\cite{wang2020multisimilaritylossgeneralpair}) improves performance over a shared objective. This configuration is used in all main experiments. Ablation studies comparing contrastive loss variants and prediction types (source-only, target-only, and bidirectional) is presented in~\ref{sec:ablation}.
All methods are trained on the same curated dataset introduced in SCFlow~\cite{ma2025scflow} except for the baseline encoders (\eg, CLIP\cite{li2019learning}) and pretrained methods (\eg, CSD\cite{somepalli2024CSD}, DEADiff\cite{Qi_2024}).
Unless otherwise indicated, we encode data using a frozen CLIP ViT-L/14 encoder and operate entirely in its embedding space. Additional experiments are conducted in ALIGN~\cite{jia2021scalingvisualvisionlanguagerepresentation} and DINOv2~\cite{dino} spaces.

\subsubsection{Datasets and Evaluation Splits}
\label{sec:exp_splits}
We evaluate under three complementary setups that progressively emphasize compositional generalization and robustness. 
(i) \textbf{Original SCFlow test set:}3,000 unseen content labels combined with the 51 training styles, measuring generalization to novel content while keeping the style set fixed~\cite{ma2025scflow}.
(ii) \textbf{In-domain unseen-style test set:} 14 additional styles excluded from training but curated using the same pipeline, evaluating style generalization within the synthetic domain. Detailed definitions and splits are provided in the Supplementary Material.
(iii) \textbf{Out-of-domain evaluation:} ImageNet~\cite{deng2009imagenet}, WikiArt~\cite{saleh2015large}, DomainNet~\cite{domainnet}, and DTD~\cite{dtd}, which differ significantly in source, structure, and visual statistics, assessing robustness under real-world distribution shift.

\subsubsection{Baselines}
\label{sec:exp_baselines}
We compare against both discriminative and generative baselines.
For discriminative baselines, we train separate content and style encoders with the same architecture and training data, using standard deep metric learning (DML) objectives: Contrastive Loss~\cite{contrastiveloss}, InfoNCE~\cite{infoNCE}, Margin Loss~\cite{wu2018samplingmattersdeepembedding}, Multi-Similarity Loss~\cite{wang2020multisimilaritylossgeneralpair}, and Proxy Anchor Loss~\cite{kim2020proxyanchorlossdeep}.
We also include CLIP~\cite{radford2021learning} and CSD~\cite{somepalli2024CSD} as strong contrastive references.
For generative baselines, we compare to DEADiff~\cite{Qi_2024} and SCFlow~\cite{ma2025scflow}, which are trained purely with generative objectives (diffusion or flow matching).

\subsubsection{Evaluation Metrics}
We evaluate disentanglement in terms of semantic correctness, factor separation, and robustness under distribution shift.
Classical disentanglement metrics such as DCI~\cite{eastwood2018a}, SAP~\cite{kumar2018variationalinferencedisentangledlatent}, and MIG~\cite{chen2019isolatingsourcesdisentanglementvariational} assume axis-aligned latent factors and are therefore ill-suited for high-dimensional embedding spaces where content and style are distributed across correlated dimensions.
Instead, we rely on retrieval-, similarity-based, and cluster-quality metrics that directly probe the semantic structure of content and style embeddings.

\vspace{0.25em}
\noindent\textbf{(i) Semantic correctness.}
We measure how well the learned embeddings match their intended semantics using zero-shot retrieval accuracy with text prompts (\Cref{tab:zeroshot}) and cosine similarity~\cite{hessel2021clipscore} between predicted embeddings and text descriptions for both reverse (see the Supplementary Material) and forward inference (\Cref{tab:fid+cossim}). 

\noindent\textbf{(ii) Disentanglement and separation.}
We quantify intra-factor clustering quality by computing normalized mutual information (NMI)~\cite{manning2009NMI} after $k$-means clustering and comparing clusters to ground-truth content and style labels (\Cref{tab:seen-nmi+sil}) alongside silhouette scores~\cite{silhouette} capturing clusters separation in the latent space (\Cref{tab:seen-nmi+sil}).

\noindent\textbf{(iii) Generation quality for forward inference.}
We use FID~\cite{heusel2018ganstrainedtimescaleupdate} with CLIP features for forward merging to measure how well the generated samples align with real data (\Cref{tab:fid+cossim}).

\noindent\textbf{(iv) Open-set and real-world robustness.}
For in-domain unseen styles, we report F1@$k$~\cite{van1974foundation} and the snapping rate (\Cref{tab:snap}). %
For out-of-domain datasets, we report rank-based retrieval scores (\Cref{tab:real-f1}). %
Finally, we valuate open-set classification on WikiArt with OSCR~\cite{openauc} curves and AUOSCR~\cite{openauc}  (\Cref{fig-oscr-wikiart}).

\subsection{Quantitative Experiments}

\subsubsection{Semantic Correctness}
\label{sec:style/content correctness}
We evaluate whether our improved method enhances output correctness in both merging (forward inference) and separation (reverse inference). For reverse inference, we assess the predicted embeddings using text prompts derived from the content and style descriptions in our dataset. As shown in~\Cref{tab:zeroshot}, vanilla SCFlow performs notably worse than CLIP in zero-shot retrieval, particularly for content.
For forward inference,~\Cref{tab:fid+cossim} reports cosine similarity with respect to content and style descriptions alongside FID scores~\cite{heusel2018ganstrainedtimescaleupdate}. Since both SCFlow and our method operate in the CLIP embedding space, we use the CLIP encoder as the feature extractor for FID computation. Across these metrics, our method surpasses CLIP in content retrieval and achieves substantial improvements in style accuracy. Additionally, forward outputs demonstrate improved FID and cosine similarity compared to SCFlow, indicating enhanced semantic alignment and generation quality in both inference directions. 

\begin{table}
\renewcommand{\arraystretch}{1.3}
\begin{center}
\caption{
Zero-shot retrieval with text prompt (CLIP embedding).
}
\label{tab:zeroshot}
\begin{tabular}{ l | c | c | c }
\toprule
\textbf{Model} & \textbf{Content} & \textbf{Style} & \textbf{Unseen Style}\\
 & \textbf{Acc. $\uparrow$} & \textbf{Acc. $\uparrow$} & \textbf{Acc. $\uparrow$} \\
\hline
CLIP~\cite{radford2021learning} & 0.7522 & 0.6844 & 0.6643\\
\hline
SCFlow~\cite{ma2025scflow} & 0.4894 & 0.6016 & 0.3857\\ 
\hline
\rowcolor{gray!8}\textbf{Ours} & \textbf{0.7694} & \textbf{0.8908} & \textbf{0.7221}\\
\bottomrule 
\end{tabular}
\end{center}
\end{table}

\begin{table}
\renewcommand{\arraystretch}{1.3}
\begin{center}
\caption{
FID and cosine similarity between style/content descriptions and forward output.
}
\label{tab:fid+cossim}
\begin{tabular}{ l | c | c | c }
\toprule
\textbf{Model} & \textbf{FID $\downarrow$} & \textbf{Style Sim. $\uparrow$} & \textbf{Content Sim. $\uparrow$} \\
\hline 
SCFlow~\cite{ma2025scflow} & 25.90 & 0.2096 & 0.0905 \\
\hline
\rowcolor{gray!8}\textbf{Ours}& \textbf{15.77} & \textbf{0.2253} & \textbf{0.1064} \\
\bottomrule
\end{tabular}
\end{center}
\end{table}

\subsubsection{Disentanglement}
\label{sec:exp-disentangle}
In order to assess the separation of content and style, we follow SCFlow and compute the normalized mutual information (NMI) score~\cite{manning2009NMI} from the obtained embeddings in~\Cref{tab:seen-nmi+sil}. NMI applies K-means clustering on the embeddings and measures the correctness of class assignments and their separability. Notably, several pure contrastive losses significantly outperform vanilla SCFlow, suggesting that generative objectives alone are insufficient to achieve effective disentanglement and class separation. However, when augmenting SCFlow with contrastive loss, we achieve the best results, validating our motivation for integrating generative and discriminative training.

The separation between learned style and content representations in the latent space is measured using silhouette score~\cite{silhouette}
which is also improved by our method, demonstrating enhanced disentanglement in the latent space.
\begin{table}
\renewcommand{\arraystretch}{1.3}
\begin{center}
\caption{
Normalized mutual information of Content and Style Clusters and silhouette score (reverse inference).
}
\label{tab:seen-nmi+sil}
\begin{tabular}{ l | c | c | c }
\toprule
\multirow{2}{*} \textbf{Model} & \textbf{Content}  & \textbf{Style} & \textbf{Silhouette}\\
& \textbf{NMI}  $\uparrow$ & \ \textbf{NMI}  $\uparrow$ & \textbf{Score} $\uparrow$ \\
\hline
CL~\cite{contrastiveloss}& 0.4598 & 0.2905 & - \\
\hline
InfoNCE~\cite{infoNCE}& 0.2327 & 0.5904 & - \\ 
\hline
Margin~\cite{wu2018samplingmattersdeepembedding}& 0.9280 & 0.9011 & - \\
\hline 
MS~\cite{wang2020multisimilaritylossgeneralpair}& 0.8950 & 0.8974 & - \\
\hline 
ProxyAnchor~\cite{kim2020proxyanchorlossdeep}& 0.7127 & 0.9214  &- \\
\hline
CSD~\cite{somepalli2024CSD}& 0.1888 & 0.8229 & - \\
\hline 
CLIP~\cite{radford2021learning}& 0.3676 & 0.5838 & - \\
\hline 
\hline
DEADiff~\cite{Qi_2024}& 0.6459 & 0.5083 & - \\
\hline 
SCFlow~\cite{ma2025scflow}& 0.8098 & 0.8559 & 0.2738\\
\hline
\hline
\rowcolor{gray!8}\textbf{Ours} & \textbf{0.9551} & \textbf{0.9493} & \textbf{0.2987}\\
\bottomrule
\end{tabular}
\end{center}
\end{table}

\subsubsection{Generalization}
\label{sec:exp-generalize}
To evaluate generalization ability, which remains less explored in SCFlow, we conduct experiments on two types of datasets: \textbf{In-domain unseen styles}, which refer to 14 unseen style classes curated under the same data collection process as the training dataset. This evaluate whether the model can generalize to novel styles within the same domain. \textbf{Out-of-domain datasets}, on the other hand, refer to external datasets such as ImageNet~\cite{deng2009imagenet}, WikiArt~\cite{saleh2015large}, DomainNet~\cite{domainnet}, and DTD~\cite{dtd}, which assess the robustness of models under large distribution shifts.

\begin{table}
\renewcommand{\arraystretch}{1.3}
\begin{center}
\caption{
Open-set retrieval: 14 unseen styles as queries against a reference space of 51 seen styles. Snapping rate indicates the probability that the model assigns unseen queries to one of the seen classes.
}
\label{tab:snap}
\resizebox{0.49\textwidth}{!}{
\begin{tabular}{ l | c | c | c | c }
\toprule 
\multirow{2}{*}{\textbf{Model}} & \multicolumn{3}{c|}{\textbf{Unseen Style F1@$k$} $\uparrow$} & \textbf{Snapping} \\
 & R/P/F1@1 & F1@10 & F1@50 & \textbf{Rate}$\downarrow$\\
\hline
CL~\cite{contrastiveloss}& 31.79 & 32.82 & 25.04 & 44.21 \\ 
\hline
InfoNCE~\cite{infoNCE}& 43.00 & 48.46 & 34.22 & 36.64 \\ 
\hline
Margin~\cite{wu2018samplingmattersdeepembedding}& 50.93 & 50.58 & 36.97 & 32.07 \\
\hline 
MS~\cite{wang2020multisimilaritylossgeneralpair}& 12.14 & 18.15 & 16.49 & 16.43 \\
\hline 
ProxyAnchor~\cite{kim2020proxyanchorlossdeep}& 47.07 & 48.46 & 34.22 & 36.50 \\
\hline
CSD~\cite{somepalli2024CSD}& 43.71 & 53.05 & 49.02 & 40.29 \\
\hline 
CLIP~\cite{radford2021learning}& 38.29 & 45.40 & 42.02 & 33.93 \\
\hline 
\hline
DEADiff~\cite{Qi_2024}& 23.57 & 31.15 & 31.40 & 60.21 \\
\hline 
SCFlow~\cite{ma2025scflow}& 55.86 & 65.97 & 57.51 & 23.14 \\
\hline
\hline
\rowcolor{gray!8}\textbf{Ours} & \textbf{65.00} & \textbf{68.33} & \textbf{61.32} & \textbf{16.29} \\
\bottomrule
\end{tabular}
}
\end{center}
\end{table}

\begin{figure}[!t]
\centering
\includegraphics[width=.8\columnwidth]{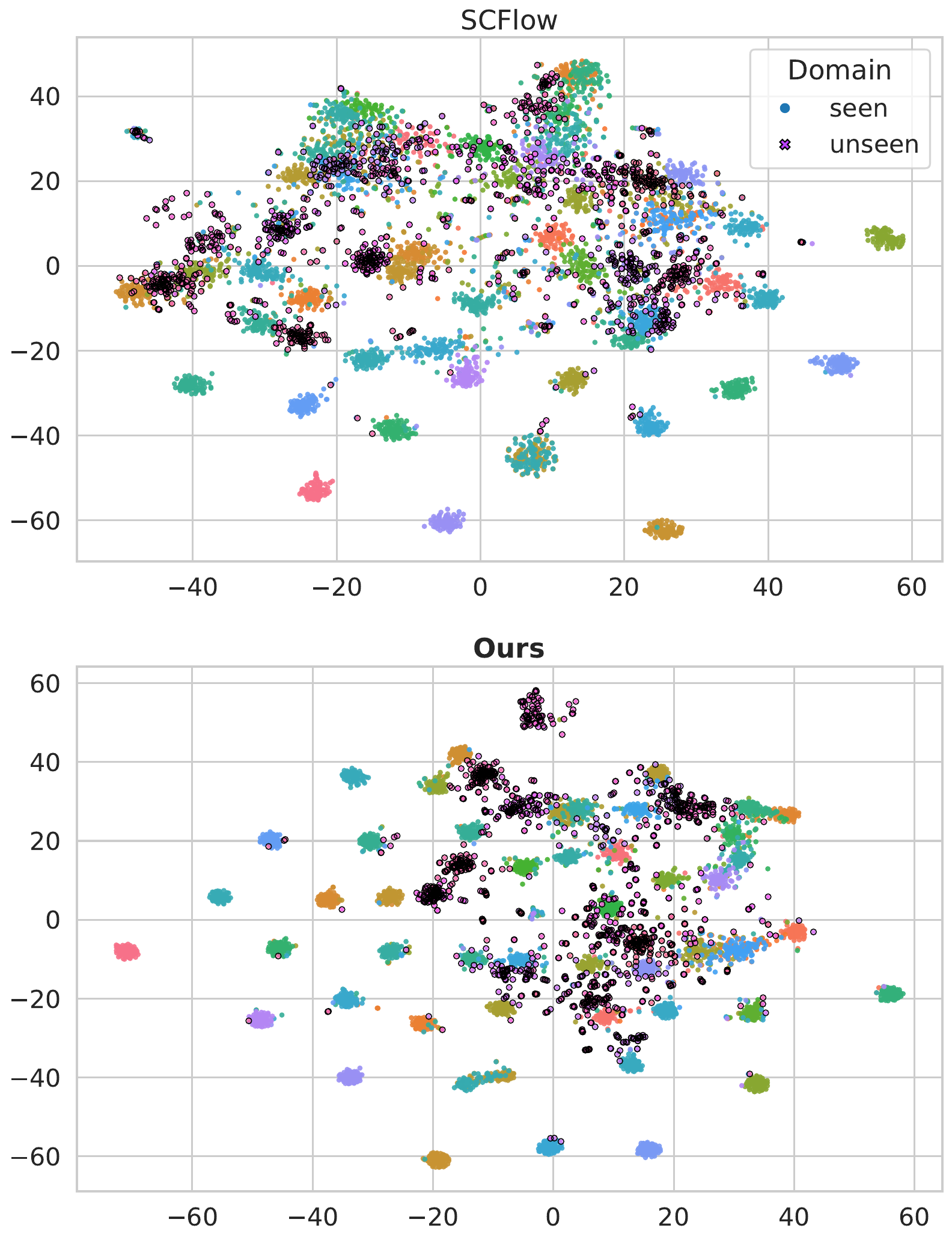}
\caption{T-SNE visualization of all 65 style classes.}
\label{fig-tsne}
\end{figure}

\noindent\textbf{In-domain Unseen Test Set:}
\Cref{tab:snap} reports F1$@k$ performance on in-domain unseen styles. The retrieval space includes both unseen and seen style embeddings: the 14 unseen styles serve as queries against all 51 seen styles from the original test set. This setup avoids the limited expressiveness of restricting retrieval to unseen styles only and provides a more realistic evaluation. The rightmost column reports the snapping rate, \ie, the probability that unseen queries are assigned to seen classes; ideally, this value should be low. Although the multi-similarity (MS) loss achieves a relatively low snapping rate, its retrieval accuracy is substantially lower, suggesting that certain purely contrastive objectives can distinguish seen from unseen samples but struggle to capture fine-grained structure within unseen classes. Our method achieves both the lowest snapping rate and the highest F1 score, indicating improved rejection of seen classes while maintaining strong retrieval among unseen styles. In addition, t-SNE visualizations in~\Cref{fig-tsne} show that our representations form more compact and better-separated clusters for both seen and unseen styles compared to SCFlow.

\noindent\textbf{Out-of-domain Test Set:}
In SCFlow, ImageNet~\cite{deng2009imagenet} and WikiArt~\cite{saleh2015large} are used to evaluate the generalization ability to unseen data. We further extend this evaluation by incorporating DomainNet~\cite{domainnet} and DTD~\cite{dtd}. DomainNet covers six domains (clipart, infograph, painting, quickdraw, real, sketch) across 345 object categories, enabling joint assessment of content and style generalization. DTD contains 47 texture classes and provides a complementary evaluation of style-focused representations.

We report F1@1 and F1@10 for similarity-based retrieval (~\Cref{tab:real-f1}). Our method achieves the best overall performance across ImageNet, WikiArt, DomainNet (Label), and DTD. While slightly below CSD and DEADiff on DomainNet style evaluation, it consistently outperforms SCFlow and all DML baselines. Notably, vanilla SCFlow remains inferior to CLIP despite being trained on CLIP embeddings, indicating that purely generative objectives underutilize the discriminative embedding space, limiting robustness and generalization.

\begin{figure}[!t]
\centering
\includegraphics[width=.8\columnwidth]{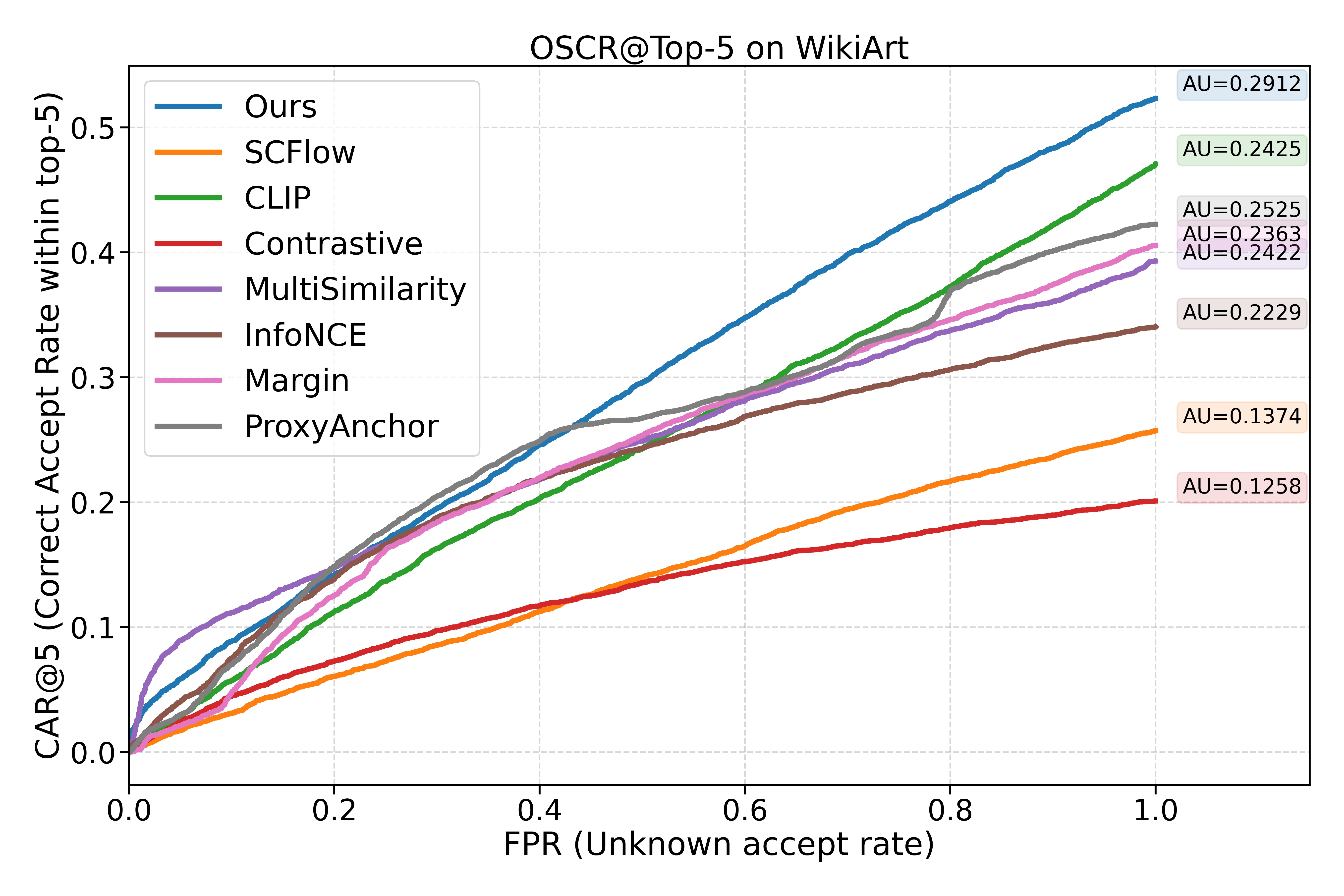}
\caption{OSCR~\cite{openauc} Curves of Classification using Wikiart Query(Top5).}
\label{fig-oscr-wikiart}
\end{figure}

We further evaluate robustness under distribution shift using open-set classification on WikiArt, following Open-AUC~\cite{openauc}. WikiArt contains 27 styles, of which 9 overlap with the 51 training styles and 18 styles are unseen. The task is to classify samples from seen styles while rejecting unseen ones, complements retrieval-based experiments by testing robustness in a classification framework. Our method achieves the highest area under the open-set classification rate (AUOSCR), with SCFlow reaching less than half of our score (~\Cref{fig-oscr-wikiart}). CLIP ranks second, likely due to large-scale pretraining. Detailed AUOSCR and NMI results are summarized in the Supplementary Material.

\noindent\textbf{Generalization to New Representation Spaces:} To assess robustness across representation spaces, we further evaluate SCFlow and our method using frozen DINOv2~\cite{dino} and ALIGN\cite{jia2021scalingvisualvisionlanguagerepresentation} encoders.
The resulting F1@K on the real-world dataset are reported in~\Cref{tab:dino_align_real_f1}. Across both spaces, our method consistently outperforms SCFlow across all datasets and surpasses the base encoder on WikiArt and DomainNet (Domain). This demonstrates that the proposed objective improves style discrimination in a representation-agnostic manner, while preserving content recognition performance.

\begin{table*}[t]
\renewcommand{\arraystretch}{1.3}
\begin{center}
\centering
\caption{Real-world Dataset Retrieval (reverse inference).}
\label{tab:real-f1}
\resizebox{0.99\textwidth}{!}{
\begin{tabular}{  l | c | c | c | c | c | c | c | c | c | c }
\toprule
\multirow{2}{*}{\textbf{Model}} & \multicolumn{2}{c|}{\textbf{ImageNet} $\uparrow$} & \multicolumn{2}{c|}{\textbf{WikiArt} $\uparrow$} & \multicolumn{2}{c|}{\textbf{DomainNet (Domain)} $\uparrow$} & \multicolumn{2}{c|}{\textbf{DomainNet (Label)} $\uparrow$} & \multicolumn{2}{c}{\textbf{DTD $\uparrow$}} \\
 & R/P/F1@1 & F1@10 & R/P/F1@1 & F1@10 & R/P/F1@1 & F1@10 & R/P/F1@1 & F1@10 & R/P/F1@1 & F1@10 \\
\hline
\hline
CL~\cite{contrastiveloss}& 56.72 & 50.43 & 59.92 & 67.37 & 83.11 & 87.25 & 49.62 & 50.12 & 59.75 & 58.20 \\
\hline
InfoNCE~\cite{infoNCE}& 63.16 & 61.18 & 65.44 & 71.82 & 86.52 & 89.76 & 55.13 & 56.60 & 69.24 & 67.93 \\ 
\hline
Margin~\cite{wu2018samplingmattersdeepembedding}& 60.02 & 60.83 & 41.81 & 50.10 & 73.28 & 80.06 & 59.12 & 62.99 & 61.79 & 59.82 \\
\hline 
MS~\cite{wang2020multisimilaritylossgeneralpair}& 60.76 & 60.11 & 48.88 & 56.23 & 77.14 & 83.06 & 59.87 & 63.08 & 40.16 & 38.09 \\
\hline 
ProxyAnchor~\cite{kim2020proxyanchorlossdeep}& 39.88 & 41.32 & 34.99 & 44.36 & 69.44 & 77.40 & 47.20 & 51.11 & 39.98 & 41.45 \\
\hline
CSD~\cite{somepalli2024CSD}& 56.55 & 59.77 & 67.35 & 72.73 & \textbf{88.47} & \textbf{91.59} & 57.68 & 60.08 & 71.10 & 69.80 \\
\hline 
CLIP~\cite{radford2021learning}& 70.14 & 71.32 & 68.11 & 73.69 & 86.64 & 89.49 & 64.42 & 68.38 & 74.73 & 75.65 \\
\hline 
\hline
DEADiff~\cite{Qi_2024}& 62.82 & 66.15 & 67.43 & 72.64 & 88.46 & 91.53 & 60.59 & 63.56 & 2.66 & 4.00 \\
\hline 
SCFlow~\cite{ma2025scflow}& 68.06 & 68.44 & 67.01 & 72.59 & 87.37 & 90.51 & 64.44 & 68.54 & 66.40 & 64.20 \\
\hline
\hline
\rowcolor{gray!8}\textbf{Ours} & \textbf{71.40} & \textbf{73.23} & \textbf{70.98} & \textbf{74.08} & 88.12 & 91.08 & \textbf{66.82} & \textbf{70.66} & \textbf{76.86} & \textbf{76.23} \\
\bottomrule
\end{tabular}
}
\end{center}
\end{table*}

\begin{table*}[t]
\renewcommand{\arraystretch}{1.3}
\begin{center}
\centering
\caption{Real-world Dataset Retrieval in DINOv2~\cite{oquab2024dinov2learningrobustvisual} and ALIGN~\cite{jia2021scalingvisualvisionlanguagerepresentation} embedding space (Reverse inference).}
\label{tab:dino_align_real_f1}
\resizebox{0.99\textwidth}{!}{
\begin{tabular}{  l | l | c | c | c | c | c | c | c | c | c | c }
\toprule
\textbf{Embedding} & \textbf{Model} & \multicolumn{2}{c|}{\textbf{ImageNet}$\uparrow$} & \multicolumn{2}{c|}{\textbf{WikiArt}$\uparrow$} & \multicolumn{2}{c|}{\textbf{DomainNet (Domain)}$\uparrow$} & \multicolumn{2}{c|}{\textbf{DomainNet (Label)}$\uparrow$} & \multicolumn{2}{c}{\textbf{DTD}$\uparrow$} \\
 \textbf{Space} & & R/P/F1@1 & F1@10 & R/P/F1@1 & F1@10 & R/P/F1@1 & F1@10 & R/P/F1@1 & F1@10 & R/P/F1@1 & F1@10 \\
\hline
\hline
DINO~\cite{oquab2024dinov2learningrobustvisual} & Raw & \textbf{76.32} & \textbf{79.57} & 51.59 & 56.00 & 81.89 & 85.23 & \textbf{61.00} & \textbf{65.42} & 75.75 & \textbf{78.99} \\
\hline
 & + SCFlow~\cite{ma2025scflow} & 75.38 & 78.67 & 52.27 & 57.53 & 83.13 & 86.32 & 60.39 & 64.79 & 75.53 & 76.66 \\
\hline
\rowcolor{gray!8}- & \textbf{+ Ours} & 75.98 & 78.81 & \textbf{57.87} & \textbf{61.84} & \textbf{84.15} & \textbf{87.59} & 60.74 & 65.34 & \textbf{77.57} & 78.35 \\
\hline
\hline
ALIGN~\cite{jia2021scalingvisualvisionlanguagerepresentation} & Raw & \textbf{62.94} & \textbf{64.68} & 59.49 & 65.46 & 88.18 & 90.66 & \textbf{60.46} & \textbf{64.35} & \textbf{74.65} & \textbf{75.86} \\
\hline
 & + SCFlow~\cite{ma2025scflow} & 59.58 & 60.82 & 59.66 & 64.85 & 87.94 & 90.82 & 52.56 & 53.47 & 69.95 & 69.04 \\
\hline
\rowcolor{gray!8}- &\textbf{+ Ours} & 62.00 & 63.81 & \textbf{63.64} & \textbf{67.31} & \textbf{88.69} & \textbf{91.51} & 59.98 & 64.11 & 72.61 & 72.99 \\
\bottomrule
\end{tabular}
}
\end{center}
\end{table*}

\subsection{Qualitative Experiments}

\begin{figure}[!t]
\centering
\includegraphics[width=\columnwidth]{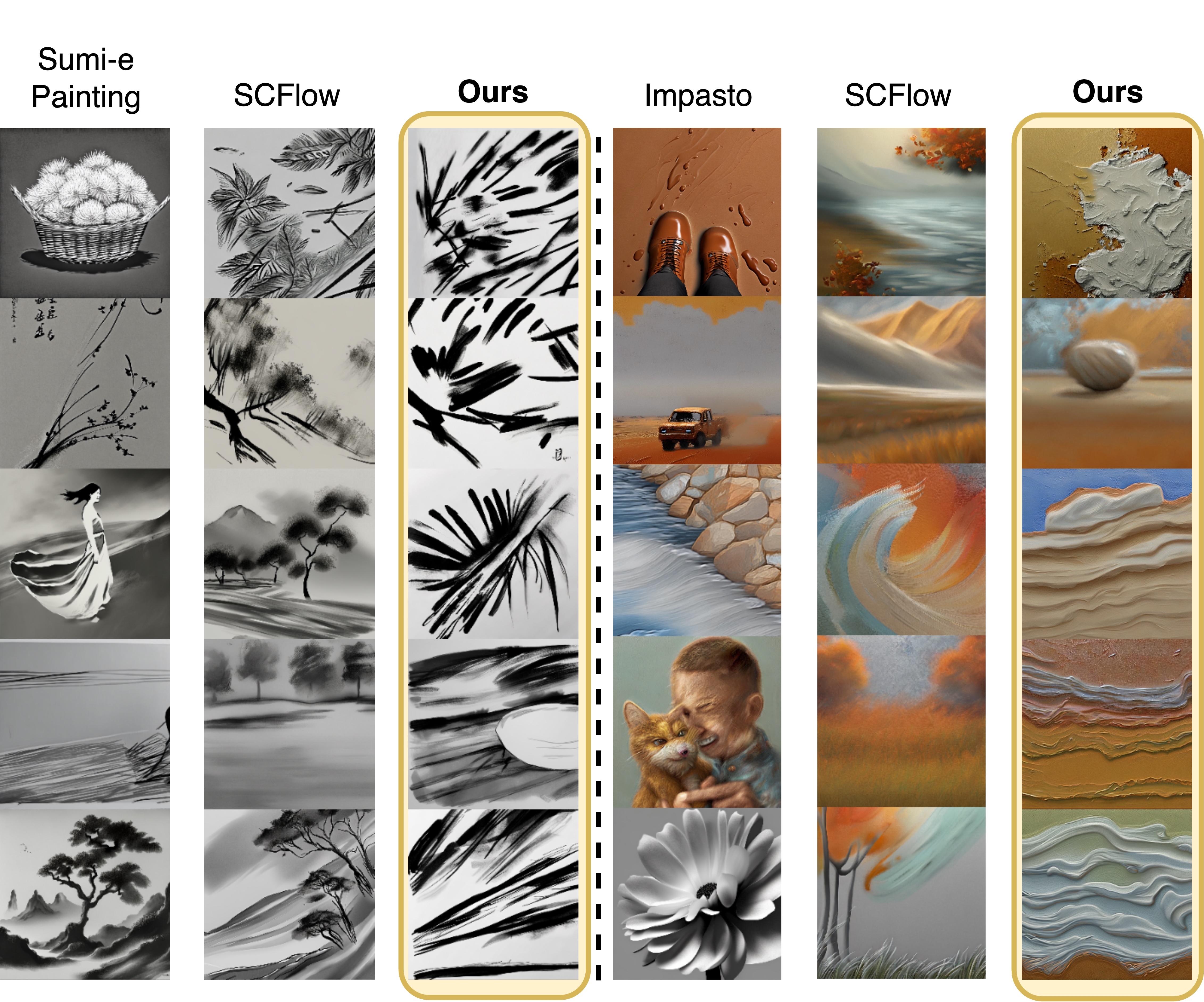}
\caption{Style Outputs. (Refer to the Supplementary Material for more)}
\label{fig-backward-style}
\end{figure}

In this section, we evaluate visual results from two complementary perspectives: the purity of style representations and the semantic fidelity of the content representations. We further assess generalization to out-of-domain data as a measure of robustness and real-world applicability.

\begin{figure}[!t]
\centering
\includegraphics[width=\columnwidth]{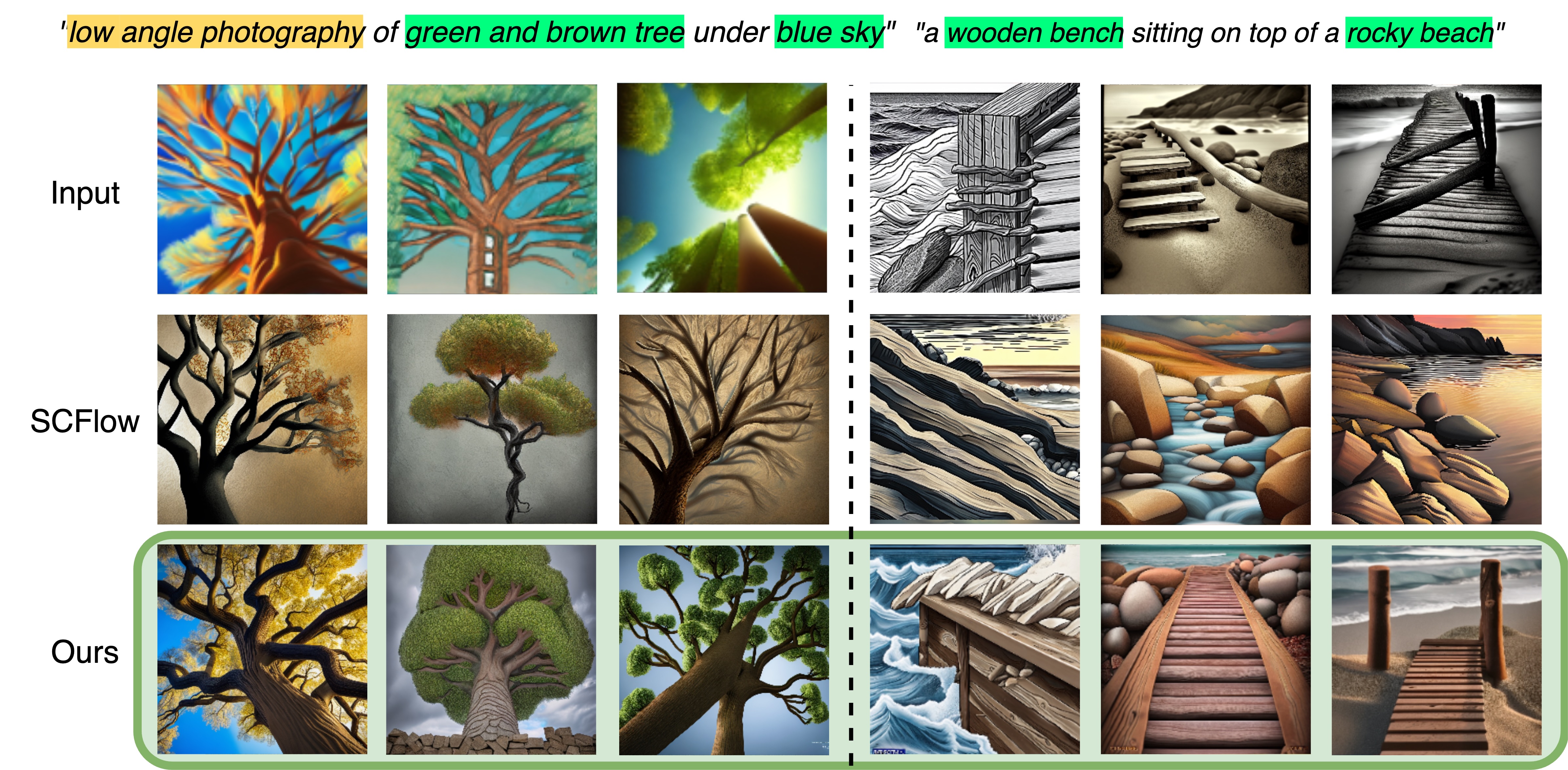}
\caption{
Content outputs.
(Refer to the Supplementary Material for more)}
\label{fig-content-cor}
\end{figure}

\subsubsection{Style Visualization}
\label{sec:style_clean}

\Cref{fig-backward-style} shows that our method produces more disentangled style representations. In contrast to SCFlow, which retains residual content structures in the extracted style embeddings, our approach isolates stylistic patterns with minimal content leakage. For instance, under the \textit{Sumi-e Painting} style, our model captures characteristic brushstroke textures and ink distributions, whereas SCFlow preserves object-level content cues in the style output.

\subsubsection{Content Transfer}
\label{sec:content_clea}
Capturing the full semantic content of an input image is critical for robust representation learning. As shown in~\Cref{fig-content-cor}, vanilla SCFlow tends to emphasize dominant objects or central regions, particularly for complex scenes containing multiple entities (\eg \textit{a wooden bench on a rocky beach}) or implicit attributes such as viewpoint (\eg \textit{low-angle photography}). In contrast, our method preserves a broader range of semantic cues, producing content representations that better reflect the complete scene context. Additional examples are provided in the Supplementary Material.

Interestingly, content outputs from the in-domain unseen test set, rendered with the same UnCLIP pipeline as SCFlow, appear noticeably more photorealistic under our model. Although photorealism is not explicitly optimized, this suggests that contrastive regularization improves semantic consistency in the learned representations. Representative examples are shown in the Supplementary Material.

\subsubsection{Real-World Domain Generalization}
\label{subsec_vis_realworld}

\begin{figure}[!t]
\centering
\includegraphics[width=\columnwidth]{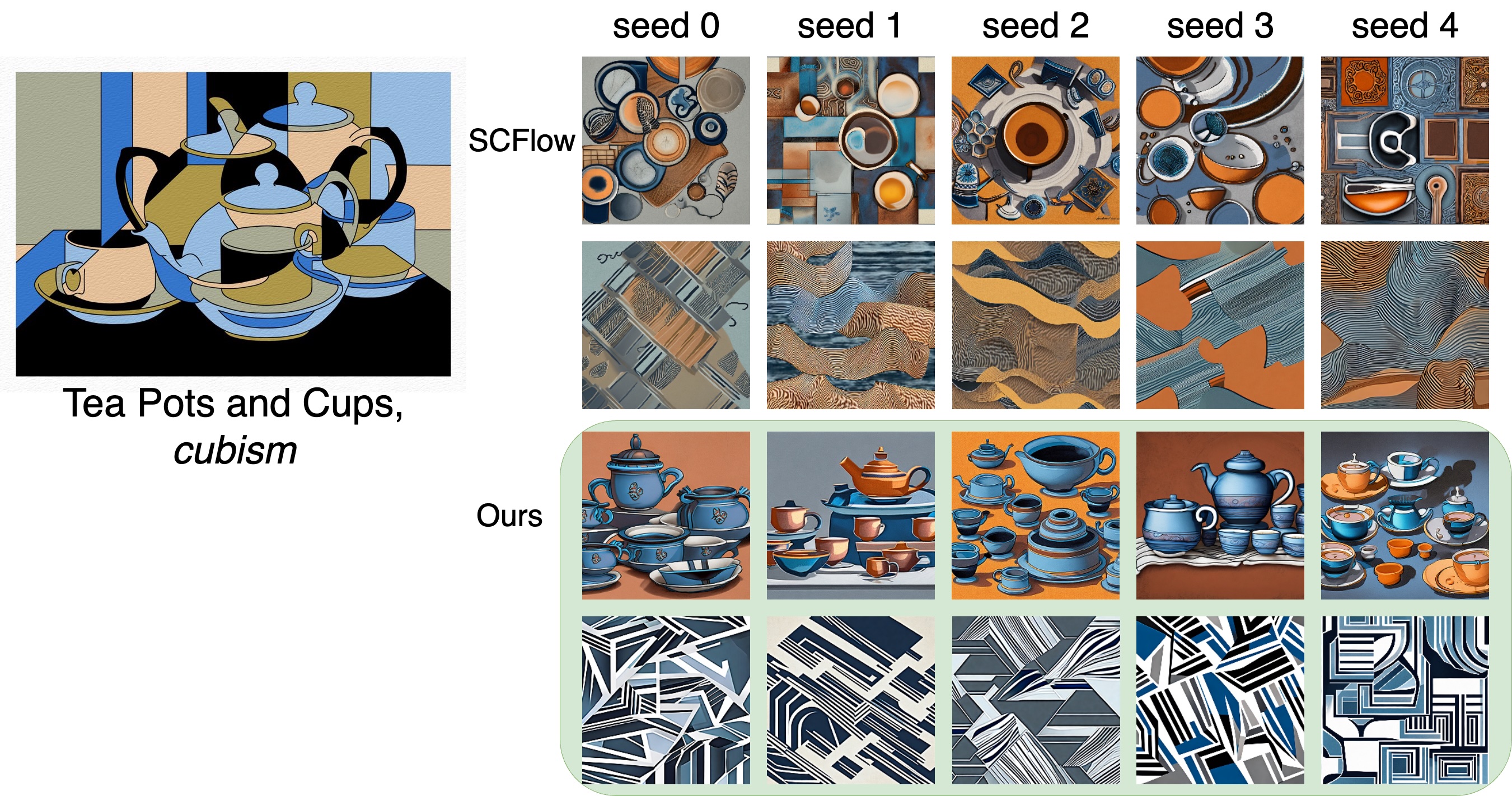}
\caption{Disentangled content and style from real-world art images. (Refer to the Supplementary Material for more.)}
\label{fig-realworld}
\end{figure}

As both SCFlow and our method are trained exclusively on synthetic data, evaluating generalization to real-world images is critical. In~\Cref{fig-realworld}, we present real-world examples together with their corresponding content and style outputs generated by both methods, using identical UnCLIP seeds~\cite{ramesh2022hierarchical} to ensure a controlled comparison. SCFlow frequently alters the input viewpoint, introduces spurious artifacts, and hallucinates additional elements not present in the original scene. Furthermore, stylistic characteristics of \textit{Cubism} are only weakly expressed in its style outputs. In contrast, our method preserves the original viewpoint, avoids extraneous artifacts, and produces style renderings that more consistently reflect the geometric abstraction and structural fragmentation characteristic of the target style.

\subsection{Ablations}
\label{sec:ablation}
We conduct ablation studies to evaluate the effect of different contrastive objectives within our framework. Since \emph{CAtFM} relies on bidirectional endpoint prediction, we also examine a unidirectional variant that predicts only the source ($\hat{x}_0$) or only the target ($\hat{x}_1$). 
We compare four contrastive losses: InfoNCE~\cite{infoNCE}, Margin Loss~\cite{wu2018samplingmattersdeepembedding}, Multi-Similarity~\cite{wang2020multisimilaritylossgeneralpair} (MS), and Proxy Anchor Loss~\cite{kim2020proxyanchorlossdeep}. Each loss is evaluated under three prediction settings: source-only, target-only, and bidirectional prediction. We analyze results along two axes: (1) comparison between purely discriminative objectives and their integration within our flow-based framework, and (2) performance differences across our prediction variants.

\begin{table*}[ht]
\renewcommand{\arraystretch}{1.3}
\begin{center}
\centering
\caption{Real-world Dataset Retrieval (Reverse inference).}
\label{tab:abl_real_f1}
\resizebox{0.9\textwidth}{!}{
\begin{tabular}{  l | l | c | c | c | c | c | c | c | c | c | c }
\toprule
\textbf{Contrastive} & \textbf{Predicted} & \multicolumn{2}{c|}{\textbf{ImageNet}$\uparrow$} & \multicolumn{2}{c|}{\textbf{WikiArt}$\uparrow$} & \multicolumn{2}{c|}{\textbf{DomainNet (Domain)}$\uparrow$} & \multicolumn{2}{c|}{\textbf{DomainNet (Label)}$\uparrow$} & \multicolumn{2}{c}{\textbf{DTD}$\uparrow$} \\
\textbf{Loss Type} & \textbf{Direction} & R/P/F1@1 & F1@10 & R/P/F1@1 & F1@10 & R/P/F1@1 & F1@10 & R/P/F1@1 & F1@10 & R/P/F1@1 & F1@10 \\
\hline
\hline
\rowcolor{gray!8}\multirow{4}{*}{InfoNCE~\cite{infoNCE}}
& - & 63.16 & 61.18 & 65.44 & 71.82 & 86.52 & 89.76 & 55.13 & 56.60 & 69.24 & 67.93 \\ 
& Source ($\hat{x}_0$) & \textbf{71.00} & \textbf{72.73} & 69.03 & \textbf{74.48} & 87.76 & \textbf{90.67} & \textbf{65.60} & 69.90 & 75.53 & \textbf{75.24} \\
& Target ($\hat{x}_1$) & 68.16 & 69.11 & 67.11 & 72.76 & 87.51 & 90.55 & 64.14 & 68.53 & 73.05 & 72.22 \\
& Bidirection (\textit{both}) & 70.46 & 72.35 & \textbf{69.20} & 74.38 & \textbf{87.85} & 90.65 & \textbf{65.60} & \textbf{70.02} & \textbf{75.62} & 74.78 \\
\hline
\hline
\rowcolor{gray!8}\multirow{4}{*}{Margin~\cite{wu2018samplingmattersdeepembedding}}
& - & 60.02 & 60.83 & 41.81 & 50.10 & 73.28 & 80.06 & 59.12 & 62.99 & 61.79 & 59.82 \\
& Source ($\hat{x}_0$) & \textbf{70.34} & \textbf{71.94} & 66.38 & 71.93 & 87.27 & 90.53 & 65.36 & \textbf{70.14} & 75.01 & 73.15 \\
& Target ($\hat{x}_1$) & 68.32 & 68.94 & 66.67 & 72.21 & 87.75 & 90.09 & 64.28 & 68.38 & 72.78 & 72.23 \\
& Bidirection (\textit{both}) & 69.64 & 70.34 & \textbf{66.74} & \textbf{72.49} & \textbf{87.89} & \textbf{91.66} & \textbf{65.77} & 69.74 & \textbf{75.18} & \textbf{74.77} \\
\hline
\hline
\rowcolor{gray!8}\multirow{4}{*}{MS~\cite{wang2020multisimilaritylossgeneralpair}}
& - & 60.76 & 60.11 & 48.88 & 56.23 & 77.14 & 83.06 & 59.87 & 63.08 & 40.16 & 38.09 \\
& Source ($\hat{x}_0$) & 69.48 & 69.96 & 60.43 & 65.98 & 85.37 & 88.61 & 64.66 & 68.82 & 73.40 & 73.73 \\
& Target ($\hat{x}_1$) & 68.84 & 69.51 & 65.76 & 71.01 & 87.57 & 90.69 & 64.34 & 68.58 & 73.05 & 72.58 \\
& Bidirection (\textit{both}) & \textbf{70.70} & \textbf{72.18} & \textbf{67.74} & \textbf{73.51} & \textbf{87.64} & \textbf{90.75} & \textbf{65.37} & \textbf{69.93} & \textbf{74.91} & \textbf{75.20} \\
\hline
\hline
\rowcolor{gray!8}\multirow{4}{*}{ProxyAnchor~\cite{kim2020proxyanchorlossdeep}}
& - & 39.88 & 41.32 & 34.99 & 44.36 & 69.44 & 77.40 & 47.20 & 51.11 & 39.98 & 41.45 \\
& Source ($\hat{x}_0$) & 67.96 & 69.75 & 61.38 & 67.27 & 84.51 & 87.98 & 63.84 & 68.38 & 68.17 & 67.40 \\
& Target ($\hat{x}_1$) & 69.00 & 69.86 & 59.38 & 65.69 & 83.50 & 87.34 & 63.38 & 67.12 & 75.09 & 74.01 \\
& Bidirection (\textit{both}) & \textbf{70.58} & \textbf{72.41} & \textbf{66.95} & \textbf{72.12} & \textbf{86.63} & \textbf{89.78} & \textbf{65.82} & \textbf{70.59} & \textbf{75.27} & \textbf{75.05} \\
\bottomrule
\end{tabular}
}
\end{center}
\end{table*}

\begin{table*}[t]
\renewcommand{\arraystretch}{1.3}
\begin{center}
\centering
\caption{Style Dataset Retrieval and NMI (Reverse inference).}
\label{tab:abl_51_recall_nmi}
\resizebox{0.85\textwidth}{!}{
\begin{tabular}{  l | l | c | c | c | c | c | c | c | c }
\toprule
\textbf{Contrastive} & \textbf{Predicted} & \multicolumn{3}{c|}{\textbf{Content F1@$k$}$\uparrow$} & \multicolumn{3}{c|}{\textbf{Style F1@$k$}$\uparrow$} & \multicolumn{2}{c}{\textbf{NMI}$\uparrow$} \\
\textbf{Loss Type} & \textbf{Direction} & R/P/F1@1 & F1@10 & F1@50 & R/P/F1@1 & F1@10 & F1@50 & Content & Style \\
\hline
\hline
\rowcolor{gray!8}\multirow{4}{*}{InfoNCE~\cite{infoNCE}}
& - & 40.84 & 31.38 & 18.65 & 68.84 & 71.59 & 68.15 & 0.2327 & 0.5904 \\ 
& Source ($\hat{x}_0$) & \textbf{81.80} & 85.82 & \textbf{77.84} & 74.20 & 85.02 & 87.95 & \textbf{0.8670} & \textbf{0.9010} \\
& Target ($\hat{x}_1$) & 79.76 & 81.49 & 72.27 & \textbf{82.36} & \textbf{87.90} & 86.03 & 0.8202 & 0.8596 \\
& Bidirection (\textit{both}) & 81.25 & \textbf{84.97} & 77.25 & 79.08 & 87.56 & \textbf{89.11} & 0.8585 & 0.8999 \\
\hline
\hline
\rowcolor{gray!8}\multirow{4}{*}{Margin~\cite{wu2018samplingmattersdeepembedding}}
& - & \textbf{94.20} & \textbf{94.27} & \textbf{85.44} & \textbf{88.33} & \textbf{90.45} & \textbf{90.56} & \textbf{0.9280} & \textbf{0.9011} \\
& Source ($\hat{x}_0$) & 85.41 & 88.29 & 79.35 & 78.90 & 87.93 & 87.53 & 0.8630 & 0.8598 \\
& Target ($\hat{x}_1$) & 82.04 & 83.64 & 72.66 & 69.00 & 79.12 & 76.66 & 0.8122 & 0.8286 \\
& Bidirection (\textit{both}) & 87.22 & 87.40 & 77.13 & 78.90 & 83.77 & 81.95 & 0.8519 & 0.8818 \\
\hline
\hline
\rowcolor{gray!8}\multirow{4}{*}{MS~\cite{wang2020multisimilaritylossgeneralpair}}
& - & \textbf{87.28} & 86.89 & 74.29 & 89.25 & 91.34 & 91.28 & \textbf{0.8950} & 0.8974 \\
& Source ($\hat{x}_0$) & 85.49 & \textbf{87.66} & 77.43 & \textbf{95.36} & \textbf{96.01} & \textbf{96.21} & 0.8460 & 0.9357 \\
& Target ($\hat{x}_1$) & 82.27 & 84.61 & 74.55 & 83.96 & 89.16 & 87.39 & 0.8257 & 0.8742 \\
& Bidirection (\textit{both}) & 79.61 & 85.72 & \textbf{77.88} & 93.08 & 95.35 & 95.49 & 0.8559 & \textbf{0.9361} \\
\hline
\hline
\rowcolor{gray!8}\multirow{4}{*}{ProxyAnchor~\cite{kim2020proxyanchorlossdeep}}
& - & 90.04 & 88.83 & 72.11 & \textbf{91.73} & 93.05 & 92.96 & 0.7127 & \textbf{0.9214} \\
& Source ($\hat{x}_0$) & 92.94 & 93.03 & 83.37 & 90.61 & 93.27 & 93.16 & 0.8876 & 0.9312 \\
& Target ($\hat{x}_1$) & 64.63 & 66.73 & 51.73 & 89.53 & 92.60 & 92.53 & 0.7387 & 0.9070 \\
& Bidirection (\textit{both}) & \textbf{95.61} & \textbf{95.49} & \textbf{87.48} & 90.96 & \textbf{93.56} & \textbf{93.79} & \textbf{0.9322} & 0.8785 \\
\bottomrule
\end{tabular}
}
\end{center}
\end{table*}

\subsubsection{Choice of Contrastive Objectives}
For the in-domain setting in~\Cref{tab:abl_51_recall_nmi}, our variants generally outperform InfoNCE, Multi-Similarity, and Proxy Anchor losses in content-embedding retrieval, while Margin loss alone achieves the strongest performance overall. A similar pattern holds for style retrieval: Margin loss performs best, and our methods surpass InfoNCE and Multi-Similarity while remaining competitive with Proxy Anchor.
These indicate that purely contrastive objectives are highly effective for similarity-based retrieval in-domain. Their disentanglement capability, reflected by higher NMI scores, can in some cases exceed ours, forming tighter clusters. 
This behavior stems from design differences: InfoNCE relies on batch-local instance contrast, making it sensitive to batch composition, whereas Proxy Anchor introduces class-level proxies that stabilize optimization. Multi-Similarity and Margin losses further enhance separation through adaptive weighting mechanisms.

Despite slightly weaker in-domain performance, our method generalizes substantially better to unseen real-world data. As shown in~\Cref{tab:abl_real_f1}, all our variants outperform purely discriminative baselines by a large margin in F1 across multiple datasets. This suggests that integrating generative and contrastive objectives yields representations that are more robust under distribution shift, whereas purely discriminative objectives tend to overfit to synthetic data despite strong within-domain results.

\subsubsection{Contrastive Guidance in Terminal Distributions}
We compare our model variants trained with different contrastive guidance strategies. As shown in~\Cref{tab:abl_51_recall_nmi}, predicting only the target endpoint $\hat{x}_1$ yields the weakest in-domain performance among the three variants, likely because the in-batch contrastive objective is applied solely to the target distribution, limiting preservation of semantic consistency along the flow trajectory.

Results in~\Cref{tab:abl_real_f1} show that the target-only variant generalizes slightly better than the source-only prediction ($\hat{x}_0$), suggesting that constraining only the source distribution may encourage mild overfitting. Given that our disentangled representations are primarily derived via reverse inference, asymmetric supervision can bias the learned structure.

Applying the contrastive objective bidirectionally aligns representations across source and target distributions, improving overall coherence. Although the source-only variant remains competitive in-domain, the bidirectional strategy consistently performs better on unseen datasets, indicating that enforcing contrastive consistency at both endpoints yields more robust and transferable representations.

%% file: Sec/Conclusion.tex
\section{Conclusion}
This work introduced \emph{CAtFM}, a framework that integrates contrastive regularization into flow matching to obtain semantically structured content and style representations. Predicting both source and target samples from the learned velocity field enables contrastive supervision on shared semantic factors, reducing latent leakage and improving representation disentanglement. To the best of our knowledge, \emph{CAtFM} is the first framework for style-content disentanglement based on flow matching that exploits bidirectional endpoint predictions as explicit learning signals.
Extensive experiments on synthetic and real-world datasets demonstrate consistent improvements over both discriminative and generative baselines. \emph{CAtFM} yields higher retrieval accuracy, clearer separation between content and style clusters, and stronger robustness under domain shift. These results underscore the advantage of coupling discriminative constraints with deterministic generative transport.

\section{Limitations and future work.}
\emph{CAtFM} currently operates in a frozen embedding space, which limits direct pixel-level controllability. An important next step is to extend the framework to VAE/diffusion latent spaces by coupling endpoint-prediction constraints with a decoder, which requires a new training pipeline for high-dimensional latents and fidelity-aware disentanglement metrics. In addition, exploring timestep-adaptive contrastive learning (\eg, hard-negative mining and schedule-aware temperature/weighting) may further reduce leakage but needs careful design to avoid shortcut solutions. A complementary theoretical direction is to analyze how endpoint-based contrastive signals modify the learned transport field and under what conditions factor-wise invariances are preserved along the flow.

%% file: Sec/Supp.tex
\section{Supplementary Material}
\label{sec:supp}

\renewcommand{\thetable}{S\arabic{table}}
\renewcommand{\thefigure}{S\arabic{figure}}

\subsection{Additional Visual Results}
\label{sec:supp-additional-visual}

\subsubsection{Latent Interpolation for Real-World Samples}
\begin{figure*}[!htb]
\centering
\includegraphics[width=.9\textwidth]{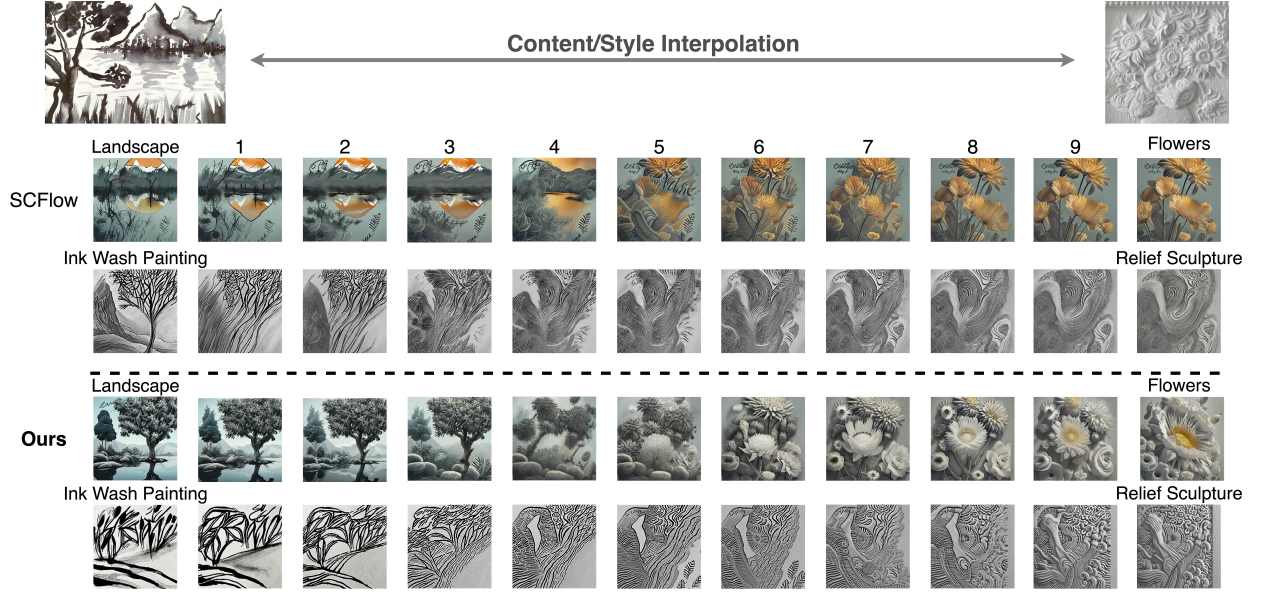}
\caption{Interpolation between obtained content and style embeddings from different real-world artistic images.}
\label{fig-interpolation}
\end{figure*}

To examine the transition behavior of our learned content and style embeddings compared with SCFlow, we visualize interpolations between embeddings derived from two real-world images, as shown in~\Cref{fig-interpolation}. In detail, each interpolation step is computed from two endpoint embeddings using the formula
\begin{equation}
    z(\lambda) = \lambda z_i + (1-\lambda)z_j, \lambda \in [0,1].
\end{equation}

Content-wise, we observe that SCFlow already begins with incomplete semantic information for the landscape example containing trees, water, and mountains. In contrast, our method produces a smoother and more coherent transition, where trees and water gradually fade, followed by the mountain, while flowers from the other endpoint slowly emerge. A similar pattern is observed in the style interpolation. The intermediate steps produced by our method vary smoothly without abrupt appearances or disappearances of style-specific features, indicating more stable and well-structured latent transitions.

\begin{figure}[!b]
\centering
\includegraphics[width=\columnwidth]{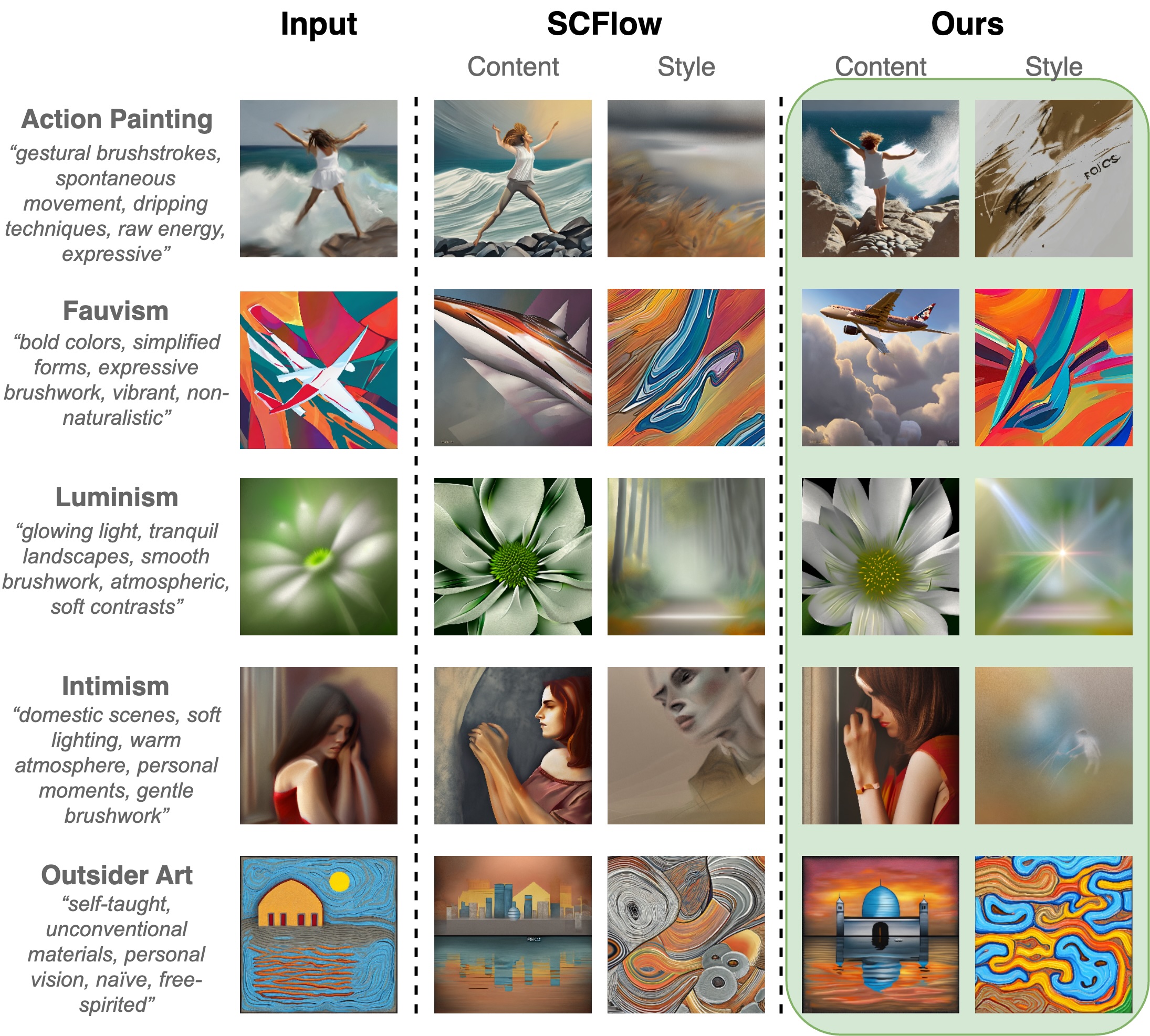}
\caption{Outputs from in-domain unseen styles along with the corresponding style descriptions. (Zoom in for better visibility)}
\label{fig-indomain-unseen}
\end{figure}

\begin{figure}[htb]
\centering
\includegraphics[width=\columnwidth]{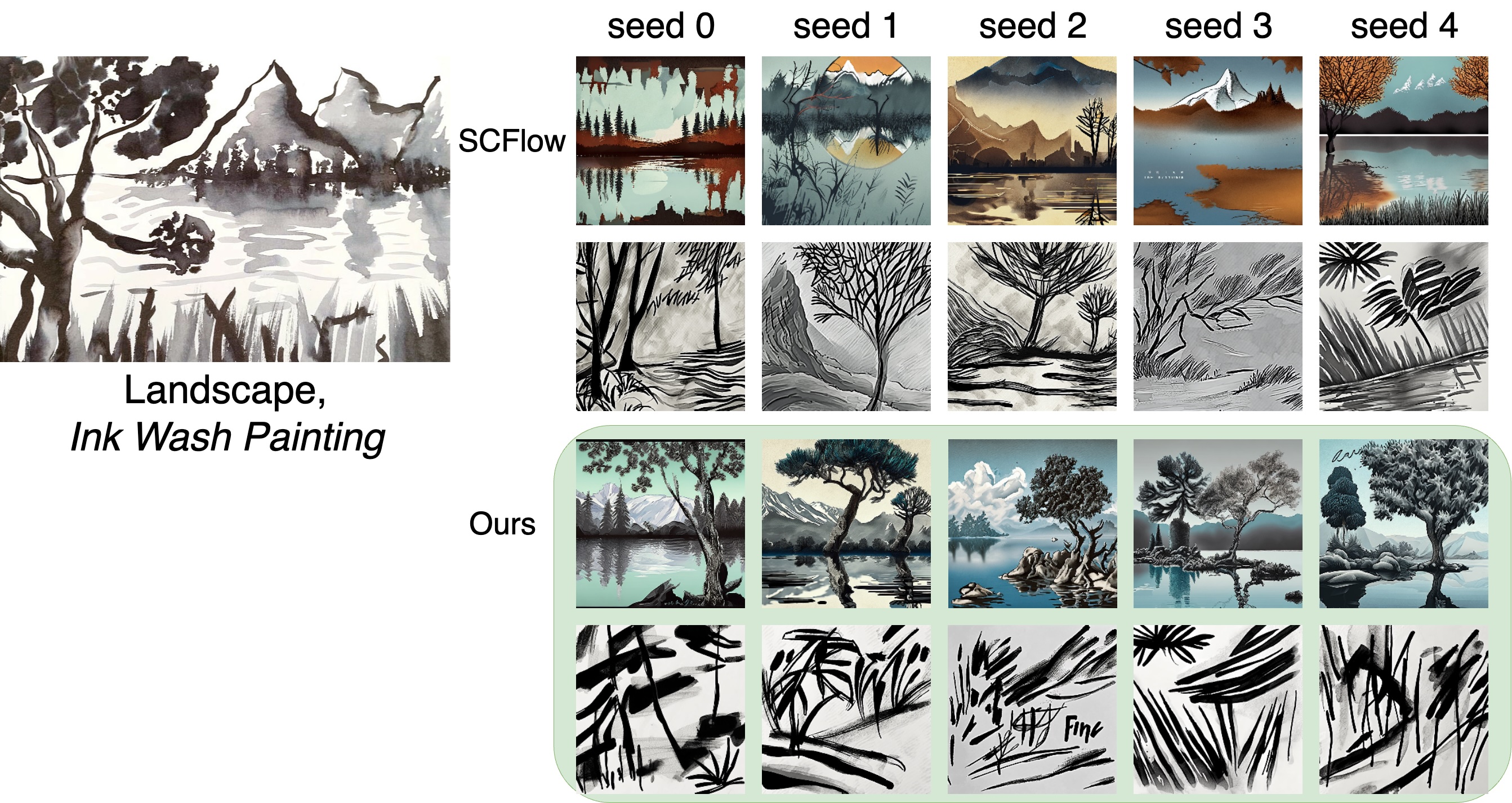}
\caption{Additional real-world visual results.}
\label{fig-real-appendix}
\end{figure}

\subsection{Extended Embedding Space}
\label{sec:supp-dino}
Aside from evaluating our method in the CLIP~\cite{radford2021learning} embedding space, we also examine its effectiveness using alternative feature encoders, such as DINOv2~\cite{oquab2024dinov2learningrobustvisual} and ALIGN~\cite{jia2021scalingvisualvisionlanguagerepresentation}. For fairness, we follow exactly the same training procedure and hyperparameters as in the main paper, replacing only the image encoder used to extract embeddings. Performance is then evaluated using the same retrieval (F1@k~\cite{van1974foundation}) and clustering metrics (NMI~\cite{manning2009NMI}).

\begin{figure*}[htb]
\centering
\includegraphics[width=.9\linewidth]{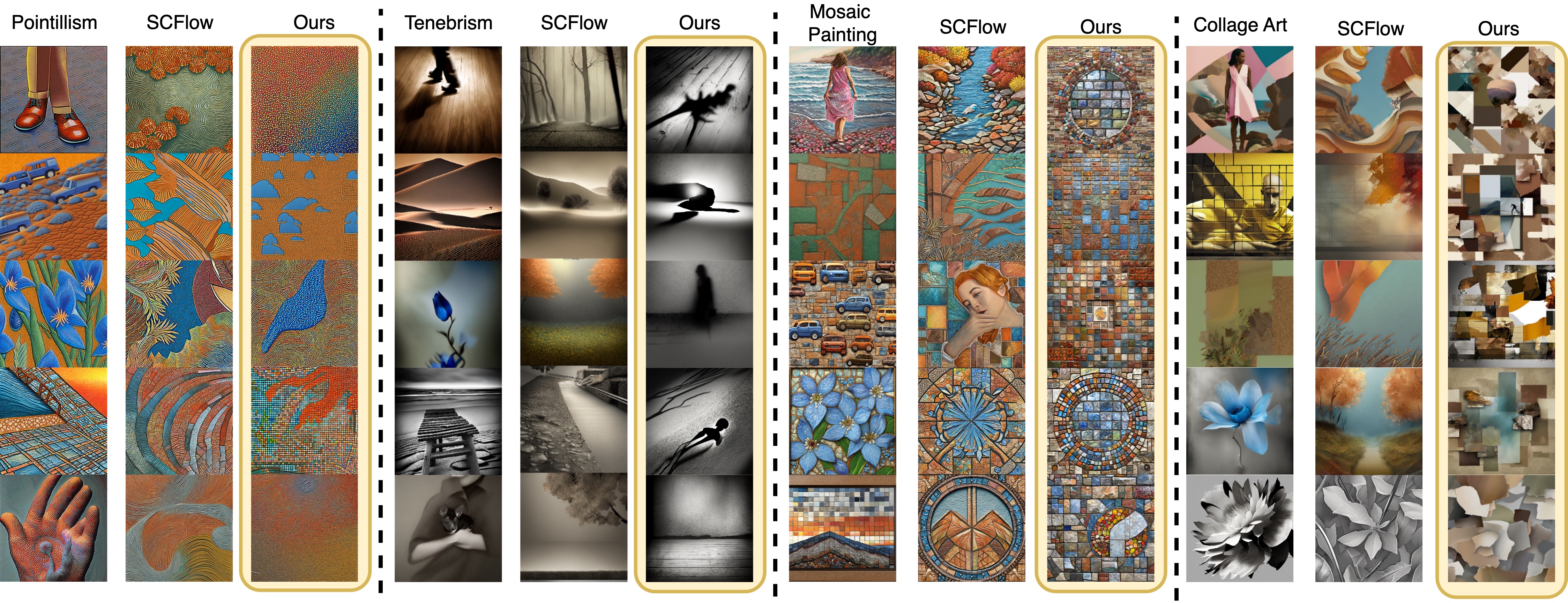}
\caption{Additional visual results for style extraction.}
\label{fig-more-styles}
\end{figure*}

\begin{figure*}[htt]
\centering
\includegraphics[width=.9\textwidth]{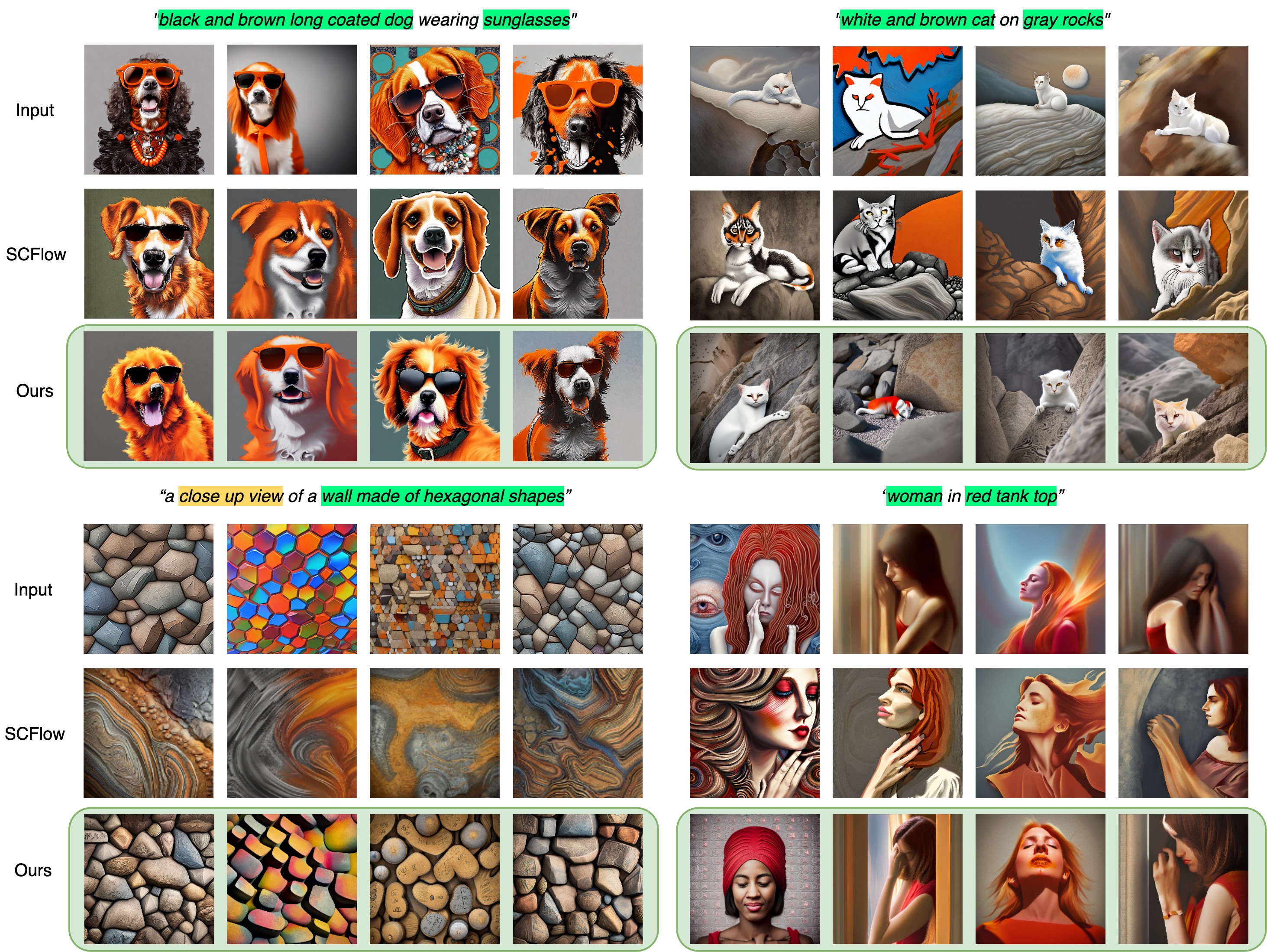}
\caption{Additional visual results for content extraction.}
\label{fig-more-contents}
\end{figure*}

As shown in~\Cref{tab:dino_51_recall_nmi}, training in the DINOv2 embedding space leads to overall improvements in both retrieval and disentanglement metrics. Across most settings, our method consistently outperforms both the raw DINOv2 features and SCFlow, with particularly large gains in style retrieval recall. In contrast, the original DINOv2 features exhibit substantially lower style recall but much higher content recall and stronger disentanglement scores, reflecting the encoder’s well-known bias toward content-oriented representations.

\begin{table*}[!ht]
\renewcommand{\arraystretch}{1.3}
\begin{center}
\centering
\caption{Style Dataset Retrieval and NMI in DINOv2~\cite{oquab2024dinov2learningrobustvisual} embedding space (Reverse inference).}
\label{tab:dino_51_recall_nmi}
\begin{tabular}{  l | l | c | c | c | c | c | c | c | c }
\toprule
\textbf{Model} & \textbf{Predicted} & \multicolumn{3}{c|}{\textbf{Content F1@$k$}$\uparrow$} & \multicolumn{3}{c|}{\textbf{Style F1@$k$}$\uparrow$} & \multicolumn{2}{c}{\textbf{NMI}$\uparrow$} \\
 & \textbf{Direction} & R/P/F1@1 & F1@10 & F1@50 & R/P/F1@1 & F1@10 & F1@50 & Content & Style \\
\hline
\hline
DINOv2~\cite{oquab2024dinov2learningrobustvisual} & - & 70.20 & 72.51 & 60.94 & 27.40 & 30.01 & 31.62 & 0.6243 & 0.1482 \\
\hline
\hline
SCFlow~\cite{ma2025scflow} & - & 77.73 & 80.26 & 68.50 & 45.40 & 54.23 & 56.21 & 0.7425 & 0.5616 \\
\hline
\hline
\rowcolor{gray!8}\multirow{4}{*}{Ours}
& Source ($\hat{x}_0$) & \textbf{82.59} & \textbf{84.84} & \textbf{74.36} & \textbf{73.04} & \textbf{83.99} & \textbf{88.84} & 0.7642 & \textbf{0.8083} \\
& Target ($\hat{x}_1$) & 78.59 & 80.40 & 68.91 & 44.44 & 52.55 & 54.63 & 0.7500 & 0.5422 \\
& Bidirection (\textit{both}) & 80.47 & 82.90 & 71.85 & 53.08 & 64.62 & 73.64 & \textbf{0.7760} & 0.7781 \\
\bottomrule
\end{tabular}
\end{center}
\end{table*}

Interestingly, the source-only ($\hat{x}_0$) variant performs slightly better than the bidirectional variant in this setting, particularly for style retrieval. Nevertheless, both our source-only and bidirectional variants consistently outperform the base DINOv2 encoder and SCFlow.

A similar trend is observed when models are trained in the ALIGN embedding space (\Cref{tab:align_51_recall_nmi}). In this case, the source-only and bidirectional variants achieve comparable performance, and both consistently surpass SCFlow as well as the raw ALIGN features.

\begin{table*}[!ht]
\renewcommand{\arraystretch}{1.3}
\begin{center}
\centering
\caption{Style Dataset Retrieval and NMI in ALIGN~\cite{jia2021scalingvisualvisionlanguagerepresentation} embedding space (Reverse inference).}
\label{tab:align_51_recall_nmi}
\begin{tabular}{  l | l | c | c | c | c | c | c | c | c }
\toprule
\textbf{Model} & \textbf{Predicted} & \multicolumn{3}{c|}{\textbf{Content F1@$k$}$\uparrow$} & \multicolumn{3}{c|}{\textbf{Style F1@$k$}$\uparrow$} & \multicolumn{2}{c}{\textbf{NMI}$\uparrow$} \\
 & \textbf{Direction} & R/P/F1@1 & F1@10 & F1@50 & R/P/F1@1 & F1@10 & F1@50 & Content & Style \\
\hline
\hline
ALIGN~\cite{oquab2024dinov2learningrobustvisual} & - & 68.55 & 68.54 & 57.37 & 38.84 & 43.31 & 46.52 & 0.6625 & 0.3464 \\
\hline
\hline
SCFlow~\cite{ma2025scflow} & - & 83.45 & 84.63 & 74.70 & 66.76 & 79.63 & 79.76 & 0.8132 & 0.8330 \\
\hline
\hline
\rowcolor{gray!8}\multirow{4}{*}{Ours}
& Source ($\hat{x}_0$) & \textbf{90.27} & \textbf{90.69} & \textbf{81.65} & \textbf{93.48} & 94.88 & 94.86 & \textbf{0.8910} & 0.9011 \\
& Target ($\hat{x}_1$) & 84.86 & 85.50 & 75.31 & 68.00 & 79.62 & 79.99 & 0.8438 & 0.8308 \\
& Bidirection (\textit{both}) & 86.04 & 89.09 & 81.24 & 93.36 & \textbf{95.38} & \textbf{96.10} & 0.8740 & \textbf{0.9360} \\
\bottomrule
\end{tabular}
\end{center}
\end{table*}

\begin{table}[h]
\renewcommand{\arraystretch}{1.3}
\begin{center}
\caption{
Our method obtains the highest normalized mutual information (NMI) scores for both unseen styles and the combined set of seen and unseen styles; it surpasses CL, InfoNCE, CLIP, CSD, and other generative baselines on silhouette scores.
}
\label{tab:NMI-unseen}
\resizebox{0.49\textwidth}{!}{
\begin{tabular}{ l | c | c | c }
\toprule
\textbf{Model} & \textbf{NMI} & \textbf{NMI} & \textbf{Sihouette} \\
& \textbf{Unseen Styles $\uparrow$} & \textbf{All Styles $\uparrow$} & \textbf{Score $\uparrow$} \\
\hline
CL~\cite{contrastiveloss}& 0.3573 & 0.2783 & 0.0090 \\
\hline
InfoNCE~\cite{infoNCE}& 0.5689 & 0.5545 & 0.1874 \\ 
\hline
Margin~\cite{wu2018samplingmattersdeepembedding}& 0.4864 & 0.7951 & 0.4242 \\
\hline 
MS~\cite{wang2020multisimilaritylossgeneralpair}& 0.2269 & 0.7801 & \textbf{0.4506}  \\
\hline 
ProxyAnchor~\cite{kim2020proxyanchorlossdeep}& 0.5314 & 0.8217 & 0.4481 \\
\hline
CSD~\cite{somepalli2024CSD}& 0.6680 & 0.6583 & 0.0849 \\
\hline 
CLIP~\cite{radford2021learning}& 0.4790 & 0.3976 & 0.0282 \\
\hline 
\hline
DEADiff~\cite{Qi_2024}& 0.5203 & 0.3632 & 0.0131 \\
\hline 
SCFlow~\cite{ma2025scflow}& 0.7264 & 0.7396 & 0.1240  \\
\hline
\hline
\rowcolor{gray!8}\textbf{Ours} & \textbf{0.9024} & \textbf{0.8618} & 0.3739 \\
\bottomrule 
\end{tabular}
}
\end{center}
\end{table}

\subsection{Dataset Details}
\label{sec:supp-data-details}

We follow SCFlow \cite{ma2025scflow} to construct the training and evaluation data using 51 artistic styles and 10,000 content categories. Content images are sourced from Pexels, and missing or sparse captions are refined using LLaVA-1.5 \cite{liu2024improved}. Style categories are curated with brief textual descriptions with ChatGPT-4o~\cite{openai2024gpt4o}. The stylized images are generated with ControlNet \cite{zhang2023controlNet}, conditioning on scribbles and using prompts of the form:

\textit{“An image depicting \{content\_caption\}, in the style of \{style\_prompt\}”}

\begin{figure}[hbt]
\centering
\includegraphics[width=.9\columnwidth]{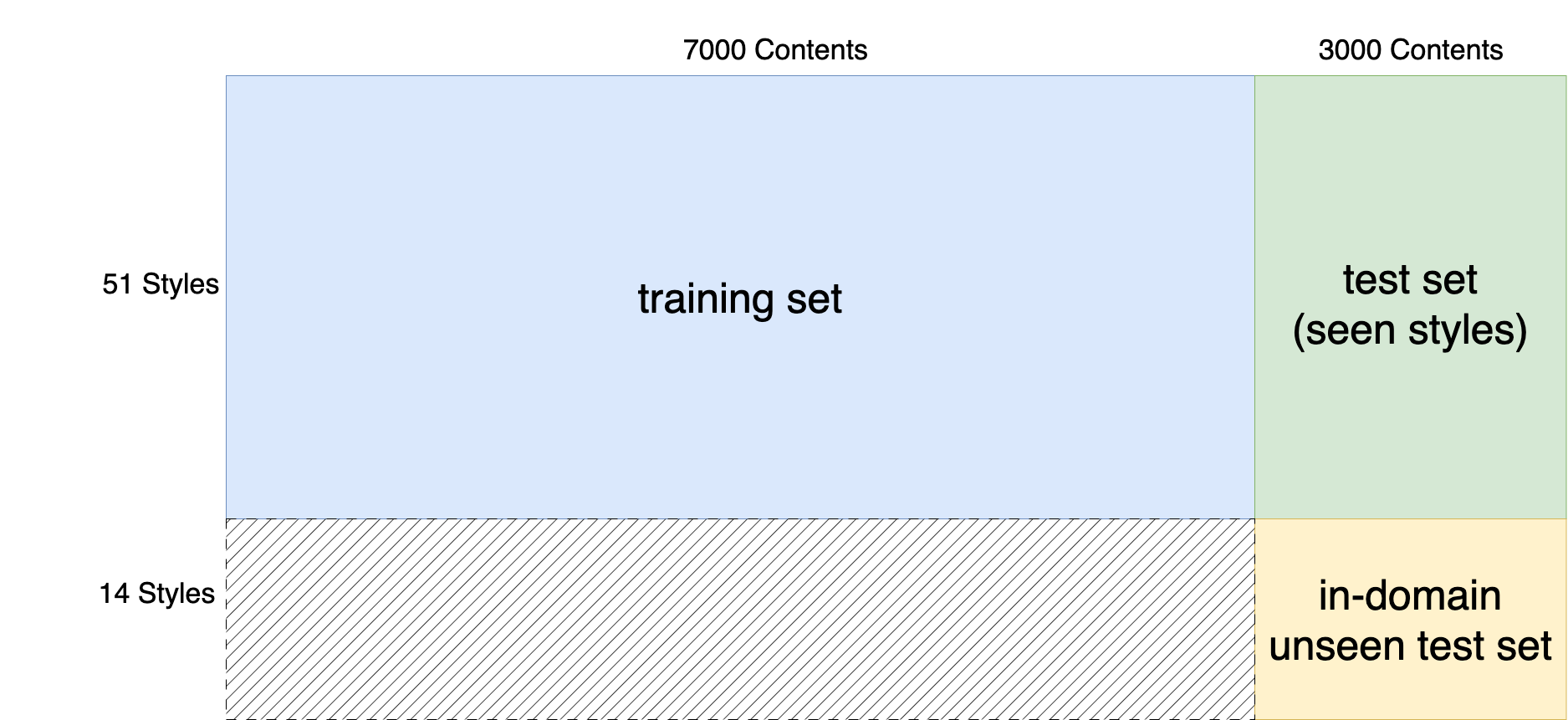}
\includegraphics[width=.9\columnwidth]{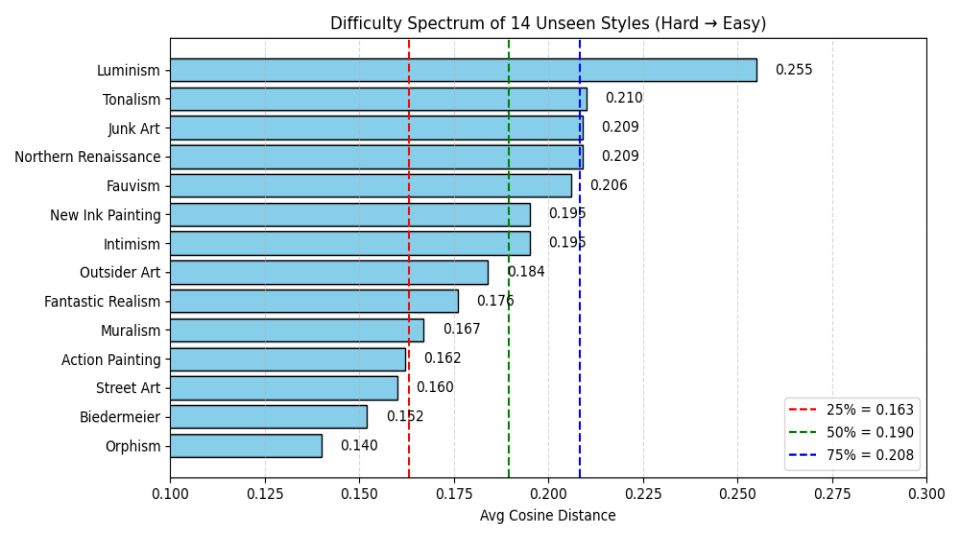}
\caption{Our new data splits and the 14 style classes from the in-domain unseen set and the corresponding spectrum of difficulties.}
\label{fig-14-unseen-style}
\label{fig-datasplit}

\end{figure}

For the in-domain unseen test set, we identify 14 additional artistic styles not covered in the original dataset and curate them using the same procedure as above. The class names are shown in~\Cref{fig-14-unseen-style} together with the corresponding average cosine distance between each unseen style’s cluster centroid and the centroids of all 51 seen styles in the CLIP embedding space. An overview of all data splits is provided in~\Cref{fig-datasplit}.

\subsection{Computational Overhead}
\label{supp_sec:compute}

\emph{CAtFM} introduces two additional components compared to SCFlow: bidirectional endpoint prediction and the incorporation of DML objectives. In principle, these additions could increase computational cost during training. To quantify the overhead, we compare the training time and GPU memory usage between the original SCFlow implementation and our method under identical training settings. Empirically, the additional cost is negligible. With a batch size of 384 for the flow-matching objective and 768 samples for the DML objective, our method incurs only 0.14\% additional training time compared to SCFlow, while GPU memory usage increases by approximately 0.1\%.

This small overhead arises because the additional objectives operate only on the final embeddings and reuse the same backbone forward pass. Since the transformer backbone dominates the overall computational cost, the extra losses introduce minimal additional computation. As a result, the proposed extensions improve representation learning while introducing virtually no additional training overhead.

\begin{table}
\renewcommand{\arraystretch}{1.0}
\begin{center}
\caption{ 
Content and Style Retrieval of Original Testset (reverse inference). Our method achieves the best style retrieval performance among all baselines. Margin Loss~\cite{wu2018samplingmattersdeepembedding} and Proxy Anchor Loss~\cite{kim2020proxyanchorlossdeep} obtain slightly higher content retrieval scores, while our model shows stronger generalization in other experiments.
}
\label{tab:51 recall}
\resizebox{0.49\textwidth}{!}{
\begin{tabular}{  l | c | c | c | c | c | c }
\toprule
\multirow{2}{*}{\textbf{Model}} & \multicolumn{3}{c|}{\textbf{Content F1@$k$}$\uparrow$} & \multicolumn{3}{c}{\textbf{Style F1@$k$}$\uparrow$} \\
 & R/P/F1@1 & F1@10 & F1@50 & R/P/F1@1 & F1@10 & F1@50 \\
\hline
\hline
CL~\cite{contrastiveloss}& 45.10 & 38.70 & 24.64 & 36.94 & 36.44 & 28.25 \\
\hline
InfoNCE~\cite{infoNCE}& 40.84 & 31.38 & 18.65 & 68.84 & 71.59 & 68.15\\ 
\hline
Margin~\cite{wu2018samplingmattersdeepembedding}& \textbf{94.20} & \textbf{94.27} & \textbf{85.44} & 88.33 & 90.45 & 90.56\\
\hline 
MS~\cite{wang2020multisimilaritylossgeneralpair}& 87.28 & 86.89 & 74.29 & 89.25 & 91.34 & 91.28 \\
\hline 
ProxyAnchor~\cite{kim2020proxyanchorlossdeep}& 90.04 & 88.83 & 72.11 & 91.73 & 93.05 & 92.96\\
\hline
CSD~\cite{somepalli2024CSD}& 49.10 & 55.22 & 52.21 & 57.39 & 66.93 & 64.09 \\
\hline 
CLIP~\cite{radford2021learning}& 58.90 & 57.86 & 46.80 & 39.61 & 30.43 & 36.86 \\
\hline
\hline
DEADiff~\cite{Qi_2024}& 61.21 & 60.22 & 49.35 & 33.14 & 40.45 & 39.73 \\
\hline 
SCFlow~\cite{ma2025scflow}& 80.78 & 83.24 & 72.68 & 80.72 & 87.46 & 85.76 \\
\hline
\hline
\rowcolor{gray!8}\textbf{Ours}& 88.27 & 87.66 & 77.43 & \textbf{93.16} & \textbf{95.35} & \textbf{95.49} \\
\bottomrule
\end{tabular}
}
\end{center}
\end{table}

\begin{table}
\renewcommand{\arraystretch}{1.3}
\begin{center}
\caption{
Normalized mutual information of WikiArt Clusters and AUOSCR (reverse inference).
}
\label{tab:wikiart}
\begin{tabular}{ l | c | c }
\toprule
\textbf{Model} & \textbf{NMI $\uparrow$} & \textbf{AUOSCR $\uparrow$} \\
\hline
CL~\cite{contrastiveloss}& 0.3399 & 0.1258 \\
\hline
InfoNCE~\cite{infoNCE}& 0.3736 & 0.2229 \\ 
\hline
Margin~\cite{wu2018samplingmattersdeepembedding}& 0.1954 & 0.2363 \\
\hline 
MS~\cite{wang2020multisimilaritylossgeneralpair}& 0.2148 & 0.2422 \\
\hline 
ProxyAnchor~\cite{kim2020proxyanchorlossdeep}& 0.1937 & 0.2526 \\
\hline
CSD~\cite{somepalli2024CSD}& 0.3464 & 0.1845 \\
\hline 
CLIP~\cite{radford2021learning}& 0.3751 & 0.2425 \\
\hline 
\hline
DEADiff~\cite{Qi_2024}& 0.3396 &  0.2199\\
\hline 
SCFlow~\cite{ma2025scflow}& 0.3644 & 0.1374 \\
\hline
\hline
\rowcolor{gray!8}\textbf{Ours} & \textbf{0.4042} & \textbf{0.2912} \\
\bottomrule
\end{tabular}
\end{center}
\end{table}

\begin{table}[htb]
\renewcommand{\arraystretch}{1.3}
\begin{center}
\caption{ Both our obtained content and style embeddings align closely with their respective text descriptions, with style embeddings showing minimal similarity to content descriptions.
}
\label{tab:cossim}
\begin{tabular}{ l | c | c | c }
\toprule
\multirow{2}{*}{\textbf{Model}} & \textbf{Content} & \textbf{Style} & \textbf{Style vs}\\
 & \textbf{Sim. $\uparrow$} & \textbf{Sim. $\uparrow$} & \textbf{Content Sim. $\downarrow$} \\
\hline
CLIP~\cite{radford2021learning} & 0.2066 & 0.2243 & -- \\
\hline
SCFlow~\cite{ma2025scflow} & 0.1825 & 0.2421 & 0.1399 \\ 
\hline
\rowcolor{gray!8}\textbf{Ours} & \textbf{0.2329} & \textbf{0.2907} & \textbf{0.0626} \\
\bottomrule 
\end{tabular}
\end{center}
\end{table}

\begin{figure}[htb]
\centering
\includegraphics[width=.9\columnwidth]{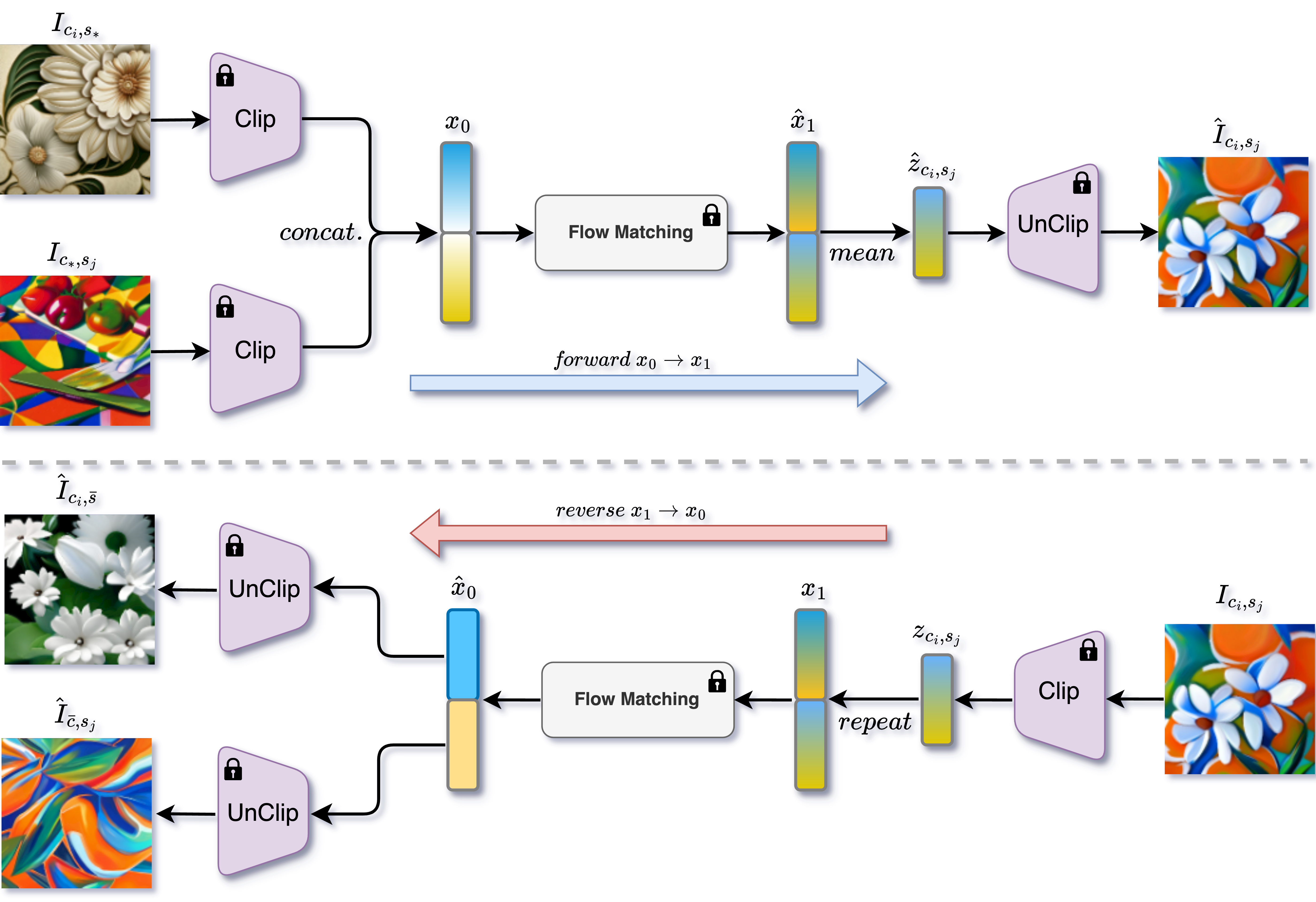}
\caption{Inference pipeline.}
\label{fig-inference}
\end{figure}